\documentclass[runningheads,table,xcdraw,dvipsnames]{llncs}

\usepackage{Styles/eccv}

\usepackage{Styles/eccvabbrv}

\usepackage{graphicx}
\usepackage{booktabs}
\usepackage{adjustbox}
\usepackage{array}
\usepackage{multirow}
\usepackage{pifont}
\usepackage{ntheorem}
\usepackage{svg}
\usepackage{algorithm}
\usepackage{algorithmic}
\usepackage{xcolor}

\usepackage[accsupp]{axessibility}  

\usepackage{hyperref}

\usepackage{orcidlink}

\newcommand{\xx}{\mathbf{x}}

\newcommand{\ww}{\mathbf{w}}
\newcommand{\calT}{\mathcal{T}}
\newcommand{\calX}{\mathcal{X}}
\newcommand{\calY}{\mathcal{Y}}
\newcommand{\calF}{\mathcal{F}}
\newcommand{\calL}{\mathcal{L}}

\newcommand{\calS}{\mathcal{S}}

\definecolor{lightred}{HTML}{fff3f3}

\begin{document}


\title{Select and Distill: \\ Selective Dual-Teacher Knowledge Transfer for Continual Learning on Vision-Language Models} 

\titlerunning{Select and Destill}

\author{
Yu-Chu Yu\inst{1, \dagger} \and
Chi-Pin Huang \inst{1} \and
Jr-Jen Chen \inst{1} \and
Kai-Po Chang \inst{1} \and
Yung-Hsuan Lai \inst{1} \and
Fu-En Yang \inst{2} \and
Yu-Chiang Frank Wang \inst{1,2,\ddagger}
}

\authorrunning{Yu et al.}


\institute{
National Taiwan University \and NVIDIA \\ $^\dagger$ \email{r09922104@ntu.edu.tw}, $^\ddagger$ \email{frankwang@nvidia.com}
}

\maketitle

\begin{abstract}
    Large-scale vision-language models (VLMs) have shown a strong zero-shot generalization capability on unseen-domain data. However, adapting pre-trained VLMs to a sequence of downstream tasks often leads to the forgetting of previously learned knowledge and a reduction in zero-shot classification performance. To tackle this problem, we propose a unique Selective Dual-Teacher Knowledge Transfer framework that leverages the most recent fine-tuned and the original pre-trained VLMs as dual teachers to preserve the previously learned knowledge and zero-shot capabilities, respectively. With only access to an unlabeled reference dataset, our proposed framework performs a selective knowledge distillation mechanism by measuring the feature discrepancy from the dual-teacher VLMs. Consequently, our selective dual-teacher knowledge distillation mitigates catastrophic forgetting of previously learned knowledge while preserving the zero-shot capabilities of pre-trained VLMs. Extensive experiments on benchmark datasets demonstrate that our framework is favorable against state-of-the-art continual learning approaches for preventing catastrophic forgetting and zero-shot degradation. Project page: \url{https://chuyu.org/research/snd}.
\end{abstract}




\section{Introduction}
\label{sec:intro}

With the access to large-scale data available for training, vision-language models (VLMs) have demonstrated unprecedented progress in visual and linguistic applications~\cite{xu2015show, antol2015vqa, zellers2019recognition, parelli2023clip}. Despite the significant achievement in static benchmark datasets, it is not easy to have VLMs incrementally accumulate the knowledge learned from previous tasks, while maintaining sufficient generalization ability. The former is known as the \emph{catastrophic forgetting} problem~\cite{mccloskey1989catastrophic}, while the latter is the zero-shot transfer capability of VLMs. 

Continual learning (CL) has emerged as a potential approach, which aims to gradually adapt the trained model to a new task without forgetting the previously learned ability. With the goal of preventing severe overfitting on currently available data and the consequent performance degradation on previous tasks, previous CL works~\cite{chaudhry2019tiny, rebuffi2017icarl, hou2019learning, lopez2017gradient, chaudhry2018efficient, riemer2018learning} are proposed to store previous data in a memory buffer. While effective for mitigating the catastrophic forgetting issue of past tasks, the scalability is hampered due to the limited memory size, restricting the deployment in the scenarios of growing fast new data. Instead of storing previous datasets in the memory buffer, recent methods~\cite{yin2020dreaming, smith2021always, gao2022r, pourkeshavarzi2021looking, choi2021dual, liu2022few} adopt a \emph{data-free} manner, which synthesizes the data of the previously trained tasks from the corresponding semantic labels. 
However, these methods are primarily designed for \emph{close-set} image recognition tasks that the label space is manually pre-determined. It remains challenging to handle the \emph{open-vocabulary} nature in vision-language models (e.g., CLIP~\cite{radford2021learning}) for zero-shot classification capability.


To address the degradation of zero-shot capabilities during sequential model fine-tuning, very recent work ZSCL~\cite{zheng2023preventing} is proposed to regularize the optimization on the current task through guidance from the original pre-trained VLMs (e.g., CLIP~\cite{radford2021learning}).
More specifically, ZSCL~\cite{zheng2023preventing} distills the knowledge from a teacher VLM, which remains frozen without fine-tuning, to constrain the fine-tuned student VLM using an \emph{unlabeled} reference dataset (e.g., ImageNet~\cite{deng2009imagenet}). This approach allows the student VLM to preserve the intrinsic zero-shot transfer capability of VLMs during fine-tuning without requiring the assessment of the pre-trained dataset. However, such a manner solely considers the zero-shot capability of VLMs. The knowledge learned from previous tasks cannot be readily preserved since the pre-trained model has never fine-tuned on previous tasks, resulting in limited performance improvement against catastrophic forgetting. Therefore, incrementally expanding the learned capability from previous tasks remains a challenging and unsolved problem.

\begin{figure}[t]
    \centering
    \includegraphics[width=\linewidth]{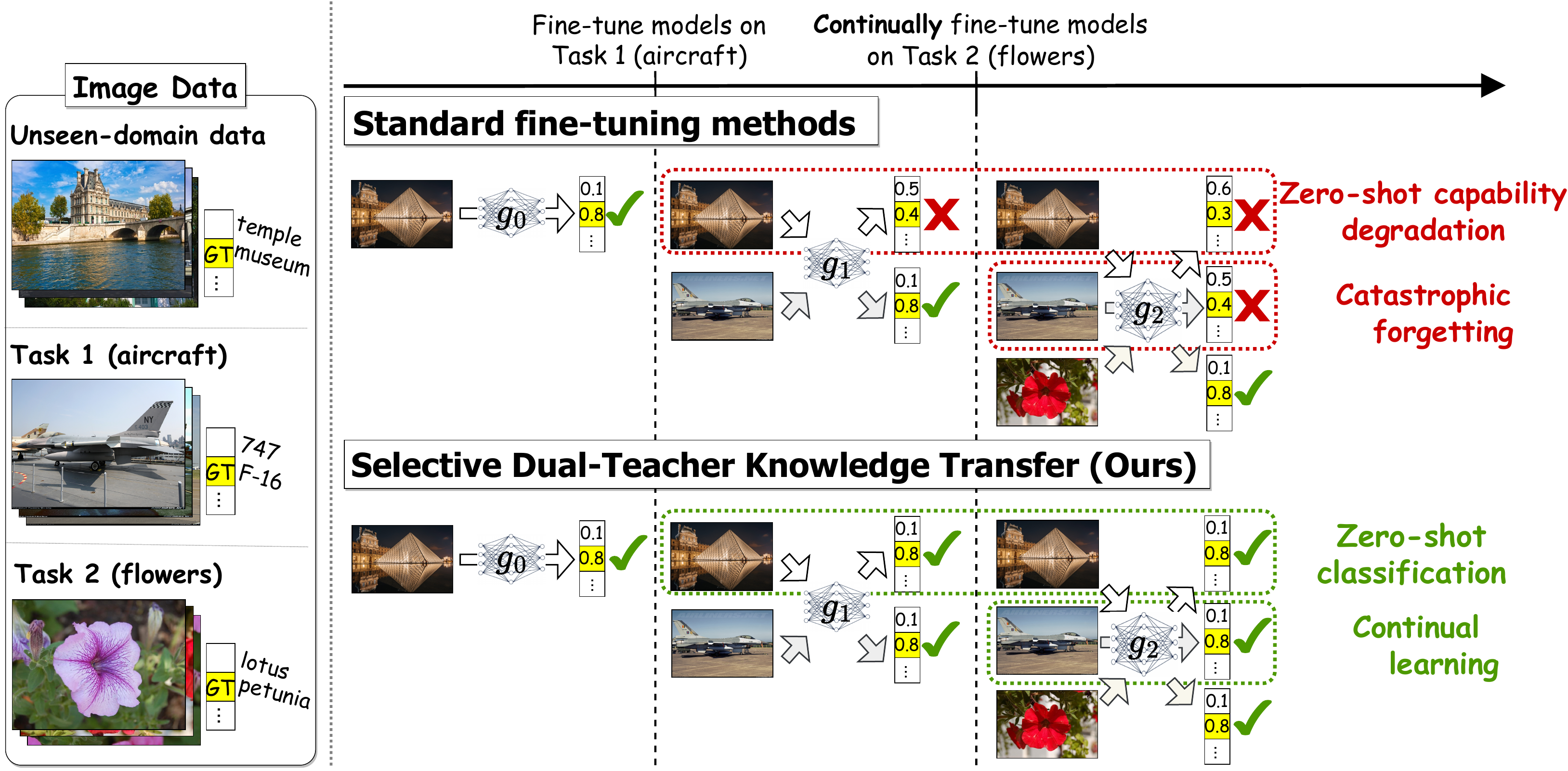}

    \caption{Compared with standard fine-tuning models, our \emph{Selective Dual-Teacher Knowledge Transfer} advances continual learning to mitigate catastrophic forgetting on previously fine-tuned tasks, while preserving the model's zero-shot capability.}
    \label{fig:teaser}
\end{figure}


In this paper, we propose a \emph{Selective Dual-Teacher Knowledge Transfer} framework, as depicted in \cref{fig:teaser}. Aiming at simultaneously enabling continual adaptation for sequentially arrived tasks while retaining the robust zero-shot transfer capability inherent in pre-trained VLMs, we follow the setting of~\cite{zheng2023preventing} and leverage both the most recent fine-tuned and the original pre-trained VLMs. Without accessing the information from previous tasks, we propose a teacher selection mechanism from dual-teacher discrepancy to identify which teacher network is favored with a given image sampled from an unlabeled reference dataset. More specifically, if the reference image aligns with prior data distribution, the most recent fine-tuned VLM would be preferable to retain the knowledge learned from past tasks. On the other hand, for other reference images that are out of the previous distribution, the original pre-trained VLM is selected to prevent the degradation of zero-shot capabilities. As a result, a selective knowledge distillation from the dual-teacher VLMs could be properly performed to enable continual learning on vision-language models.

We now summarize our contributions as below:
\begin{itemize}

\item We propose a \emph{Selective Dual-Teacher Knowledge Transfer} framework that simultaneously alleviates catastrophic forgetting problems and preserves the zero-shot capabilities from the pre-trained VLM.

\item By observing an unlabeled reference dataset, our framework views pre-trained and the most recently finetuned models as dual-teachers, and selects the proper one for knowledge distillation based on the introduced discrepancy.

\item Extensive evaluations on several benchmark datasets in various incremental learning settings confirm that our approach performs favorably against existing continual learning methods, alleviating both catastrophic forgetting and zero-shot degradation.
\end{itemize}





\section{Related Work}
\label{sec:related}


\subsubsection{Rehearsal-Based Continual Learning.} Rehearsal-based continual learning~\cite{chaudhry2019tiny, rebuffi2017icarl, hou2019learning, lopez2017gradient, chaudhry2018efficient, riemer2018learning} mitigates catastrophic forgetting by maintaining a subset of previous training data in a memory buffer, and the stored data can then be combined with current data for regularizing model fine-tuning. For example, iCaRL~\cite{rebuffi2017icarl} efficiently selects representative samples from previous tasks to maintain an evenly distributed memory buffer. LUCIR~\cite{hou2019learning} addresses the issue where the model incorrectly favors newer classes caused by potential data imbalance that exists between previous and new tasks. However, retaining data from previous tasks in a memory buffer poses risks of privacy leakage and a costly storage burden, restricting the scalability in real-world deployments.

\subsubsection{Data-Free Continual Learning.} Data-Free Continual Learning (DF-CL)~\cite{li2017learning, yin2020dreaming, smith2021always, gao2022r, pourkeshavarzi2021looking, choi2021dual, liu2022few} aims to preserve knowledge learned from past tasks without accessing their data. Several DF-CL methods~\cite{yin2020dreaming, smith2021always, gao2022r, pourkeshavarzi2021looking, choi2021dual, liu2022few} are learned to synthesize prior data given its corresponding semantic label. Then, they could regularize the fine-tuning of the current task by distilling the knowledge from previous models using the synthetic data of prior tasks. In addition, another line of methods~\cite{wang2022learning, wang2022dualprompt, wang2023attriclip} focuses on learning lightweight prompts to encode task-specific information on top of a frozen pre-trained Vision Transformer (ViT). In this way, they are able to guide the frozen pre-trained ViT to perform the current task without forgetting the previous knowledge captured in the task-specific prompts.
Although these methods have shown remarkable abilities in recalling previously learned data, they mainly focus on \emph{close-set} image classification tasks, so they still cannot readily be applied to \textit{open-vocabulary} VLMs, which require simultaneously retaining the knowledge learned from prior tasks and zero-shot capability inherent in large-scale pre-trained VLMs.

\subsubsection{Continual Learning on Vision-Language Models.} 
Recently, VLMs~\cite{radford2021learning, jia2021scaling, pham2023combined} pre-trained on large-scale datasets have demonstrated robust zero-shot transferability for open-vocabulary downstream tasks. 
However, recent studies~\cite{kumar2022fine, wortsman2022robust} have shown that the zero-shot capability is prone to deteriorate when fine-tuning the pre-trained VLMs to specific domains.
With the aim of preserving the zero-shot capability during model fine-tuning, ZSCL~\cite{zheng2023preventing} is proposed to regularize the model via the guidance from original pre-trained VLMs. Without the need to access the pre-trained dataset, ZSCL~\cite{zheng2023preventing} claims that performing knowledge distillation from pre-trained VLMs on an \emph{unlabeled} reference dataset (e.g., ImageNet~\cite{deng2009imagenet}) can effectively preserve the zero-shot capabilities during model fine-tuning. While promising, ZSCL~\cite{zheng2023preventing} primarily considers preventing zero-shot transfer degradation. It cannot easily expand the knowledge derived from sequentially arrived downstream tasks where only an unlabeled reference dataset is accessible. In this work, we propose a unique \emph{Selective Dual-Teacher Knowledge Transfer} framework, which aims at simultaneously preserving zero-shot transferability while mitigating catastrophic forgetting for previous tasks.

\section{Method}
\label{sec:method}

\subsection{Problem Formulation}
\label{subsec:preliminary}

For the sake of completeness, we first define the problem setting in this paper. In the context of continual learning, we assume that the model has been trained on $K$ sequentially arrived tasks $\{\calT^1, \calT^2, \cdots, \calT^K\}$, where the $k$-th task $\calT^k = (\calX^k, \calY^k)$ contains $N^k$ images $\calX^k = \{\xx^k_i\}_{i=1}^{N^k}$ with $L^k$ class labels $\calY^k \subseteq \{1, 2, \cdots, L^k\}$. Following~\cite{zheng2023preventing}, we only have access to the most recent fine-tuned VLM ($g_{k-1}$) and the original pre-trained VLM ($g_0$), but \textit{not} the data from previous tasks (i.e., $\{\calT^j\}_{j=1}^{k-1}$). On the other hand, an unlabeled reference dataset $\calX^{\text{ref}}$ (e.g., ImageNet~\cite{deng2009imagenet}) can be utilized to during continual learning (as~\cite{zheng2023preventing} did). For continual leaning on VLMs, the model $g_k$ is expected to preserve not only the knowledge learned from previous tasks $\{\calT^j\}_{j=1}^{k-1}$), but also the zero-shot transfer capability of large-scale pre-trained VLMs like CLIP~\cite{radford2021learning} during model fine-tuning on $\calT^k$.

\begin{figure}[t]
    \centering
    \includegraphics[width=0.8\linewidth]{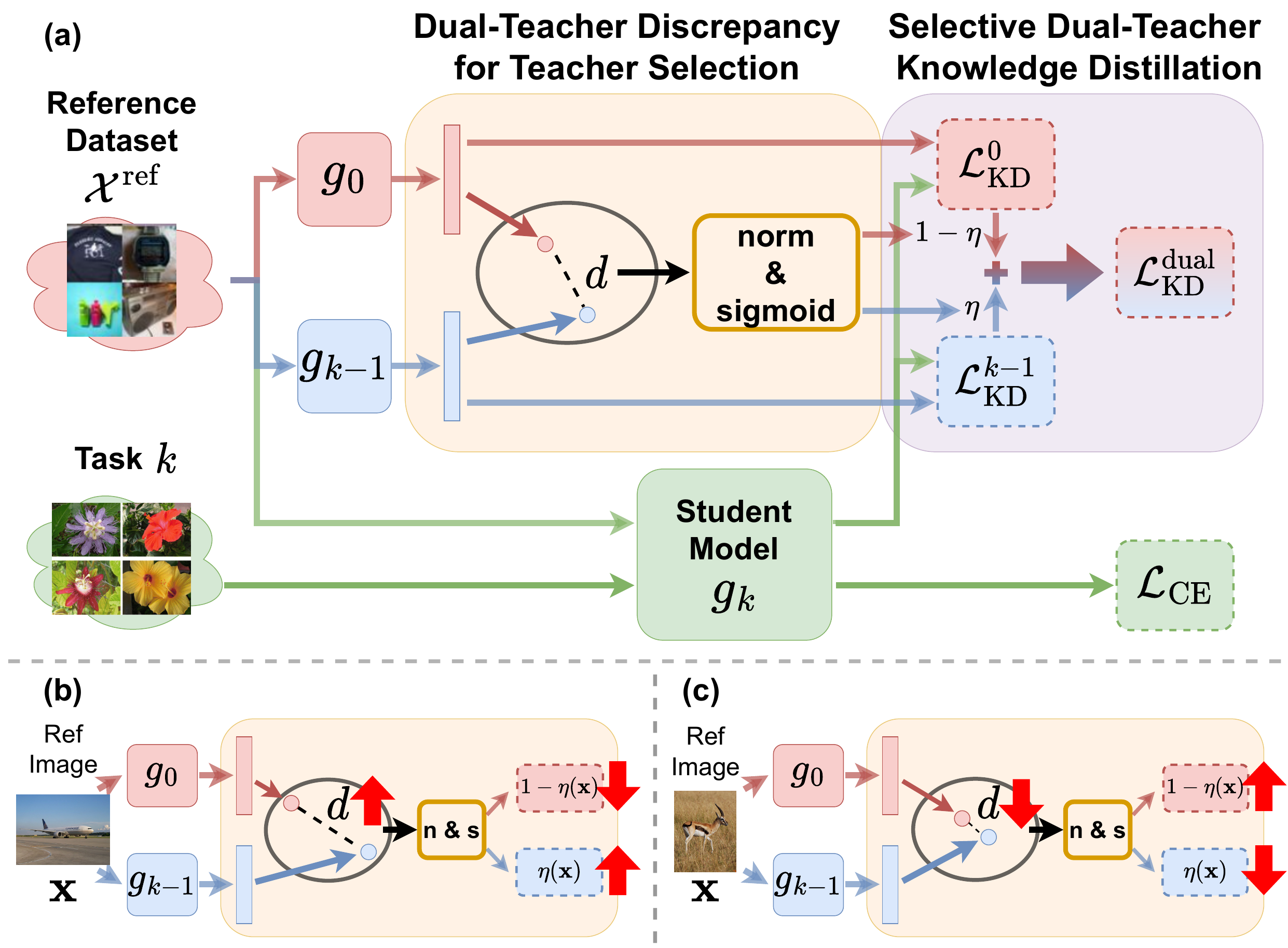}
    \caption{\textbf{(a)} The overall architecture of our proposed \emph{Selective Dual-Teacher Knowledge Transfer} framework. \textbf{(b)} Selective knowledge transfer from $g_{k-1}$ due to larger discrepancy $d$ between dual teachers $g_0$ and $g_{k-1}$, alleviating catastrophic forgetting on Task $k-1$. \textbf{(c)} Selective knowledge transfer from $g_{0}$ due to smaller discrepancy $d$ between dual teachers $g_0$ and $g_{k-1}$, preserving the zero-shot capability of $g_0$.}
    \label{fig:main_arch}
\end{figure}

\subsection{Selective Dual-Teacher Knowledge Transfer on VLMs}
\label{subsec:main_method}

Given the most recent fine-tuned VLM $g_{k-1}$ and the original pre-trained VLM $g_0$, together with an unlabeled reference dataset $\calX^{\text{ref}}$, our goal is to tackle catastrophic forgetting and to preserve zero-shot transfer capability for continual learning of VLMs. Since the alignment between $\calX^{\text{ref}}$ and data from tasks $\{\calT^1, \calT^2, \cdots, \calT^{(k-1)}\}$ is not known, direct knowledge distillation from $g_{k-1}$ on $\calX^{\text{ref}}$ might not be desirable. 

As shown in \cref{fig:main_arch}, we propose a novel framework, \emph{Selective Dual-Teacher Knowledge Transfer}, to perform VLM continual learning, aiming to alleviate catastrophic forgetting while preserving zero-shot transferability. We view $g_{k-1}$ and ($g_0$) as \emph{dual teachers} to perform selective knowledge transfer. That is, for each image extracted from $\calX^{\text{ref}}$, we need to identify the proper teacher to perform knowledge distillation. In the following subsections, we will detail how we utilize such irrelevant/unlabeled data and present our selection process. We will explain how our selective dual-teacher knowledge transfer would jointly alleviate catastrophic forgetting on data from previous tasks while retaining the zero-shot capabilities on unseen image data.


\subsubsection{Dual-Teacher Discrepancy for Teacher Selection.} 
In our work, we distill the knowledge from the dual teacher networks of the most recent fine-tuned VLM $g_{k-1}$ and the pre-trained VLM $g_0$, as depicted in  \cref{fig:main_arch}. The problem is that, one cannot easily determine the knowledge from which teacher VLM to be distilled when observing an image $\xx^{\text{ref}}$ from $\calX^{\text{ref}}$. If $\xx^{\text{ref}}$ does not match the distribution of data observed by $g_{k-1}$, performing knowledge distillation from $g_{k-1}$ would not preserve the zero-shot classification ability. On the other hand, if $\xx^{\text{ref}}$ is visually similar to the previously fine-tuned data of $g_{k-1}$, it is not desirable to distill knowledge from $g_0$, as $g_0$ lacks specific knowledge to alleviate catastrophic forgetting on $\calT^{k-1}$.

To tackle the above challenge, we propose a teacher selection mechanism based on the \emph{dual-teacher discrepancy}. To be more precise, if a sampled reference image $\xx^{\text{ref}}$ aligns with the distribution of previous datasets, the feature derived by the $g_{k-1}$ would differ from that obtained by the pre-trained VLM $g_0$, inducing large dual teacher discrepancy $d$. On the other hand, as a reference image is out of previous data distribution, a smaller discrepancy $d$ would be expected due to this reference image being unfamiliar to both teacher models, so that such unseen-domain data can be leveraged to facilitate zero-shot preservation. Thus, we can denote the relation is formulated as follows:

\begin{equation}
\label{eq:expected_comparison}
\mathop{\mathbb{E}}\limits_{\xx \in \calX^{1:k-1}} \big[d(g_{k-1}(\xx), g_0(\xx))\big] \ge \mathop{\mathbb{E}}\limits_{\xx' \notin \calX^{1:k-1}} \big[d(g_{k-1}(\xx'), g_0(\xx'))\big],
\end{equation}

\noindent where $d: \calF \times \calF \mapsto [0, \infty)$ denotes dual-teacher discrepancy measurement, which is realized by an Euclidean distance in the feature space $\calF$. $\calX^{1:k-1} = \bigcup_{i=1}^{k-1} \calX^i$ collects all data fine-tuned before, and $\xx' \notin \calX^{1:k-1}$ represents data that has not seen by $g_{k-1}$ before. The above observation is also empirically evident in \cref{tab:avg_dist} and further analyzed in \cref{subsec:analysis}.

With the dual-teacher discrepancy $d$ derived from $g_{k-1}$ and $g_{0}$, we are able to select the favored teacher VLM for knowledge transfer given the sampled reference image $\xx^{\text{ref}}$. To be more specific, we define a \textit{selection scoring function} $\eta ( \cdot): \calX \mapsto [0, 1]$ at task $k$ that transforms the discrepancy $d$ into a selection score, as computed as follows,
\begin{equation}
\label{eq:similarity_estimation}
\eta(\xx) = \sigma(\frac{d(g_{k-1}(\xx), g_0(\xx)) - \delta}{\gamma}),
\end{equation}
where $\delta, \gamma \in \mathbb{R}$ are hyper-parameters to normalize the feature discrepancy, and $\sigma: \mathbb{R} \mapsto [0, 1]$ is a sigmoid function mapping the normalized discrepancy to a scalar score between $0$ and $1$.

We note that, a \textit{larger} selection score $\eta(\xx)$ (e.g., greater than 0.5) indicates the most recent fine-tuned VLM $g_{k-1}$ would be preferable to mitigate catastrophic forgetting on prior tasks, as depicted in \cref{fig:main_arch}(b). Conversely, when the selection score $\eta(\xx)$ is \textit{small}, the pre-trained VLM $g_0$ is expected to transfer the zero-shot capabilities, as illustrated in \cref{fig:main_arch}(c). 




\subsubsection{Selective Knowledge Distillation from Dual-Teachers.}

With the estimated teacher selection score $\eta(\xx)$ obtained, we are able to perform selective knowledge transfer from the dual teacher networks. As depicted in \cref{fig:main_arch}(a), at the current task $k$, our framework selectively distills the knowledge from $g_{k-1}$ and $g_0$ to the student VLM $g_k$ to alleviate the forgetting of previous tasks and the degradation of zero-shot capabilities through the control by the teacher selection score $\eta(\xx)$. As a selective knowledge transfer from $g_{k-1}$ is preferable when $\eta(\xx)$ is large while a knowledge distillation from $g_0$ is encouraged as relatively low $\eta(\xx)$, we compute the dual-teacher knowledge distillation objective $\calL_{\text{dual}}$ as,

\begin{equation}
\label{eq:dual}
\calL_{\text{KD}}^{\text{dual}} = \sum_{\xx \sim \calX^{\text{ref}}} \eta(\xx) \cdot \calL_{\text{KD}}^{k-1} + (1 - \eta(\xx)) \cdot \calL_{\text{KD}}^{0},
\end{equation}
where $\calL_{\text{KD}}^{k-1} = d(g_{k-1}(\xx), g_k(\xx))$ denotes a knowledge distillation objective which aligns the feature representations of the input $\xx$ with that of the most recent fine-tuned model $g_{k-1}$, and $\calL_{\text{KD}}^{0}= d(g_0(\xx), g_k(\xx))$ aims to the align the feature representation with the pre-trained model $g_0$.

Combining with the standard cross-entropy loss function $\calL_{\text{CE}}$ on the current task $\calT^k$, our proposed framework is capable of retaining both previously fine-tuned knowledge from $g_{k-1}$ and the inherent zero-shot transferability from the pre-trained VLM $g_0$ during fine-tuning on task $k$. In summary, the overall objective function of our proposed \emph{Selective Dual-Teacher Knowledge Transfer} framework is formulated as below:
\begin{equation}
\label{eq:total}
    \calL = \calL_{\text{CE}} + \calL_{\text{KD}}   ^{\text{dual}}.
\end{equation}


\subsection{Training and Inference}

\subsubsection{Training Phase.}
\label{sec:training}
Following previous settings for continual learning on Vision-Language Models~\cite{zheng2023preventing}, we fine-tune the original pre-trained model CLIP~\cite{radford2021learning} to the downstream tasks in a sequential manner.
We summarize the training algorithm in our supplementary material. Note that at each stage $k$, we do not have access to data from previous tasks $\{\calT^1, \cdots, \calT^{k-1}\}$. After sequentially fine-tuning over all $K$ different tasks, we derive a final model $g_K$ that exhibits zero-shot classification capabilities with catastrophic forgetting suppressed.

\subsubsection{Inference Phase.}
Once the learning of the proposed framework is complete, we deploy the derived $g_K$ for performing image recognition tasks on each task $\{\calT^1, \calT^2, \cdots, \calT^K\}$. Following the inference manner proposed in CLIP~\cite{radford2021learning}. Let $h$ denote the text encoder of the CLIP model $g$. Given a set of labels $\calY$ with $L$ different categories, we convert the corresponding class name of each label $y$ to a text feature vector $\ww_y$. Then, the probability of a data $\xx$ to the class $y$ is calculated as below:
\begin{equation}
p(y| \xx) = \frac{\exp( \cos (g(\xx), \ww_y) / \tau)}{\sum_{j=1}^L \exp( \cos(g(\xx), \ww_j) / \tau)},
\end{equation}
where $\tau$ is a temperature parameter learned by CLIP~\cite{radford2021learning}.

\section{Experiment}
\label{sec:experiment}

\subsection{Implementation Detail}

In our experiments, we use CLIP~\cite{radford2021learning} implemented by open\_clip~\cite{ilharco_gabriel_2021_5143773} with the ViT-B/16~\cite{dosovitskiy2020image} image encoder as our backbone. We optimize our model with AdamW, dynamically adjusting learning rates with a cosine scheduler started by $1 \times 10^{-5}$ and a weight decay regularization set to $5 \times 10^{-4}$. During training, only the image encoder is updated while the text encoder is kept frozen. We standardize the text prompt to "a photo of a <CLASS>" during both the training and testing phases for classification purposes. 

\subsection{Datasets}
We evaluate our proposed method on eight fine-grained classification datasets, including FGVC-Aircraft~\cite{maji2013fine}, DTD~\cite{cimpoi2014describing}, EuroSAT~\cite{helber2019eurosat}, Flowers-102~\cite{nilsback2008automated}, Food-101~\cite{bossard2014food}, Oxford-Pets~\cite{parkhi2012cats}, Stanford-Cars \cite{krause20133d}, and UCF-101~\cite{soomro2012ucf101}. Note that to avoid potential overlapping label spaces among different datasets, we alleviate coarse-grained datasets such as Caltech-101~\cite{fei2004learning}, CIFAR100~\cite{krizhevsky2009learning}, and SUN397~\cite{xiao2010sun} used in previous works~\cite{zheng2023preventing}.

\subsection{Evaluation Protocol}
\label{subsec:evaluation}
\subsubsection{Multiple Training Sequences.}
Unlike previous benchmarks~\cite{zheng2023preventing}, which picked only one or two sequences to evaluate the performance, we construct $K$ unique sequences to fully understand the level of forgetting after multiple training rounds for each dataset. Specifically, given $K$ different image classification tasks, we first construct the first ordered sequence $\calS^1 = (\calT^1, \calT^2, \cdots, \calT^K)$. Based on $\calS^1$, we shift the sequence to the left to derive the next sequence $\calS^2 = (\calT^2, \calT^3, \cdots, \calT^K, \calT^1)$. Thus, the $k$-th sequence $\calS^k$ contains ordered tasks:
\begin{equation}
\label{eq:sequence}
\calS^k = (\calT^{k \ \%\  K}, \calT^{(k+1)\ \%\ K}, \cdots, \calT^{(k + K - 1)\ \%\ K}),
\end{equation}
where $\%$ indicates the mod operator. By training and testing on these $K$ different sequences, each dataset has one chance to be the first and the last dataset during continual training progress. This allows us to thoroughly evaluate the catastrophic forgetting and the degradation of zero-shot transferability after training on $K$ multiple rounds.

Specifically, we pick up the first training sequence $\calS^1$ in the following order: (Aircraft, DTD, EuroSAT, Flowers, Food, Pets, Cars, UCF101), and the next $7$ training sequences can be derived through \cref{eq:sequence}. We leave the detailed order of tasks for each training sequence in the supplementary.

\begin{figure}[t]
    \centering
    \includegraphics[width=.8\linewidth]{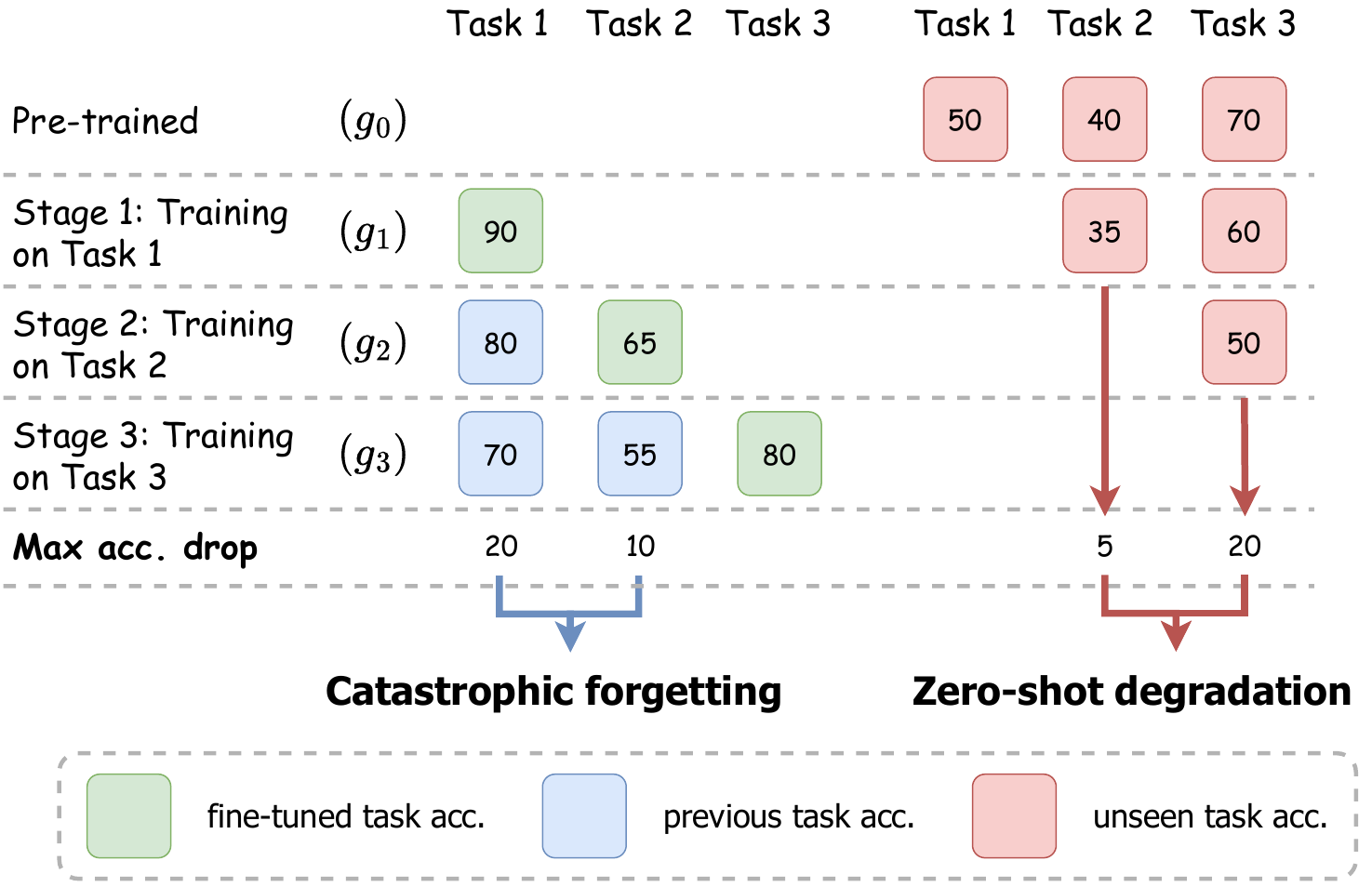}
    \caption{\textbf{Illustration of training and evaluation schemes for continual learning.} From top to bottom rows, the pre-trained model $g_0$ is incrementally finetuned on different tasks (in green). For the incrementally learned model $g$ in each row, data of unseen tasks are shown in red, while that of previously fine-tuned ones are in blue.}
    \label{fig:metrics}
\end{figure}

\subsubsection{Metrics.}

For each sequence, we measure three metrics, \emph{Average accuracy}, \emph{Catastrophic forgetting}, and \emph{Zero-shot degradation}. Following previous works~\cite{lopez2017gradient, chaudhry2018riemannian, chaudhry2019tiny}, Average accuracy is calculated as the mean value of the final performance on each task. Catastrophic forgetting measures the average of the maximum performance drop on each previous task. Also, Zero-shot degradation evaluates the average of the maximum performance drop on each unseen task. We illustrate the calculation of Catastrophic forgetting and Zero-shot degradation in \cref{fig:metrics}.


\subsection{Main Result}

\subsubsection{Baseline Methods.} We compare our proposed framework with several baseline methods. Continual FT is the most straightforward strategy that continually fine-tunes the model to the current task. LwF~\cite{li2017learning} proposes to distill knowledge from the previous model with the current data. iCaRL~\cite{rebuffi2017icarl} maintains a memory buffer that stores previous data and performs knowledge distillation to acquire knowledge from the previous models. ZSCL~\cite{zheng2023preventing} is the most related method to our approach, which also introduces a reference dataset and distills knowledge from the pre-trained CLIP model. In addition, they further apply a weight-space ensemble for certain intervals to ensure a gradual transition of model parameters. MoE-Adapters~\cite{yu2024boosting} is the newest state-of-the-art that involves incremental adapters as mixture-of-experts~\cite{jacobs1991adaptive, shazeer2017outrageously} upon a frozen CLIP model, and further apply an automatic selector to allocate data to the experts during test phase.

\begin{table}[t]
    \caption{Quantitative comparisons on Multi-Domain Task-Incremental Learning (MTIL) benchmark. In MTIL, inference is performed in a sequential manner on each dataset. $\calS^i$ denotes the $i$-th training sequence.}
    \label{tab:mtil}
    \begin{adjustbox}{width=\textwidth,center}
    
    \centering
        {\setlength{\tabcolsep}{5pt}
        \begin{tabular}{lcccccccccc}
            \bottomrule

            \toprule
             Method $/$ Sequence & $\calS^1$ & $\calS^2$ & $\calS^3$ & $\calS^4$ & $\calS^5$ & $\calS^6$ & $\calS^7$ & $\calS^8$ & \textbf{Mean}\\
            \midrule

            

            \textbf{Catastrophic forgetting ($\downarrow$)} & & & & & & & & \\

            Continual FT  & 10.98 & 10.60 & 8.80 & 19.17 & 10.11 & 11.95 & 15.19 & 9.48 &12.04\\

            LwF~\cite{li2017learning} & 10.38 & 6.52 & 6.37 & 10.22 & 7.99 & 7.70 & 10.41 & 8.91 &8.56 \\

            iCaRL~\cite{rebuffi2017icarl} & 8.42 & 7.00 & 6.45 & 10.21 & 7.03 & 7.33 & 9.68 & 8.23 &8.04 \\

            ZSCL~\cite{zheng2023preventing} & 4.67 & 2.35 & 2.13 & 2.97 & 3.15 & 4.28 & 4.89 & 4.70 & 3.64\\

            MoE-Adapters~\cite{yu2024boosting} &2.74 &4.71 &4.28 &1.15 &1.50 &1.60 &2.94 &2.77 &2.71\\

            Ours & \textbf{1.70} & \textbf{1.16} & \textbf{0.89} & \textbf{1.04} & \textbf{0.59} & \textbf{1.34} & \textbf{1.12} & \textbf{1.79} &\textbf{1.20}\\
            \midrule

            \textbf{Zero-shot degradation ($\downarrow$)} & & & & & & & & &\\

            Continual FT & 24.81 & 23.58 & 19.54 & 16.46 & 22.22 & 19.02 & 19.54 & 24.02 &21.15 \\

            LwF~\cite{li2017learning} & 10.75 & 10.23 & 8.63 & 8.25 & 12.02 & 10.33 & 8.98 & 11.01 & 10.03\\

            iCaRL~\cite{rebuffi2017icarl} & 13.77 & 12.68 & 11.28 & 12.14 & 13.20 & 13.20 & 13.09 & 14.01 & 12.92\\

            ZSCL~\cite{zheng2023preventing} & 3.44 & 3.94 & 4.02 & 2.85 & 3.79 & 2.31 &{ 1.86} &{ 1.84} & 3.00\\

            MoE-Adapters~\cite{yu2024boosting} &1.62 &2.58 &\textbf{1.04} &2.37 &4.31 &3.05 &\textbf{1.77} &\textbf{0.63} & 2.17\\
            
            Ours & \textbf{1.55} & \textbf{2.04} & {1.21} & \textbf{1.92} & \textbf{2.79} & \textbf{2.18} & {1.90} & {2.08} &\textbf{1.96} \\
            
            \midrule

            \textbf{Average accuracy ($\uparrow$)} & & & & & & & & & \\

            Continual FT & 76.16 & 76.24 & 78.03 & 68.69 & 76.64 & 75.44 & 72.71 & 77.45 &75.17 \\

            LwF~\cite{li2017learning} & 76.78 & 80.45 & 80.65 & 77.52 & 79.64 & 79.45 & 77.31 & 78.70 & 78.81 \\

            iCaRL~\cite{rebuffi2017icarl} & 77.99 & 79.77 & 79.93 & 76.66 & 79.26 & 79.08 & 77.06 & 78.61 & 78.55\\

            ZSCL~\cite{zheng2023preventing} & 81.89 & 83.98 & 84.30 & 83.49 & 83.41 & 82.38 & 81.92 & 81.97 & 82.92\\

            MoE-Adapters~\cite{yu2024boosting} & 82.71 &80.74 &81.15 &83.97 &83.68 &83.68 &82.73 &79.68 & 82.29\\

            Ours & \textbf{84.48} & \textbf{84.92} & \textbf{84.97} & \textbf{84.89} & \textbf{85.50} & \textbf{85.07} & \textbf{85.02} & \textbf{84.52} & \textbf{84.92}\\
            
            \bottomrule

            \toprule
        \end{tabular}
        }
    \end{adjustbox}
    
\end{table}

\begin{table}[t]
    \caption{Quantitative comparisons on Multi-Domain Class-Incremental Learning (MCIL) benchmark. In MCIL, the task (data domain) to be evaluated is not known during inference and thus can be viewed as \textit{open-set} classification.}
    \label{tab:mcil}
    \begin{adjustbox}{width=\textwidth,center}
    
    \centering
        {\setlength{\tabcolsep}{5pt}
        \begin{tabular}{lcccccccccc}
            \bottomrule
            
            \toprule
            Method $/$ Sequence & $\calS^1$ & $\calS^2$ & $\calS^3$ & $\calS^4$ & $\calS^5$ & $\calS^6$ & $\calS^7$ & $\calS^8$ &\textbf{Mean}\\
            \midrule

            

            \textbf{Catastrophic forgetting ($\downarrow$)} & & & & & & & & \\

            Continual FT & 11.17 & 10.89 & 10.16 & 20.12 & 10.57 & 12.14 & 15.62 & 9.80 &12.56\\
            LwF~\cite{li2017learning} & 9.56 & 6.38 & 6.93 & 11.09 & 8.37 & 7.69 & 10.24 & 8.44 &8.59\\
            iCaRL~\cite{rebuffi2017icarl} & 8.43 & 6.90 & 6.83 & 10.69 & 7.09 & 7.37 & 10.17 & 8.66 &8.27\\
            ZSCL~\cite{zheng2023preventing} & 4.21 & \textbf{1.41} & 2.08 & 3.32 & 2.85 & 4.39 & 5.22 & 5.13 &3.58\\
            Ours & \textbf{1.92} & {1.53} & \textbf{0.97} & \textbf{1.14} & \textbf{0.58} & \textbf{1.55} & \textbf{1.29} & \textbf{1.81} &\textbf{1.35}\\

            \midrule

            \textbf{Zero-shot degradation ($\downarrow$)} & & & & & & & & \\

            Continual FT & 24.54 & 24.10 & 19.53 & 17.60 & 21.96 & 18.92 & 20.26 & 24.31 &21.40\\
            LwF~\cite{li2017learning} & 11.94 & 11.82 & 8.27 & 9.99 & 13.36 & 11.47 & 10.95 & 12.63 &11.30\\
            iCaRL~\cite{rebuffi2017icarl} & 13.02 & 12.78 & 10.89 & 11.81 & 12.74 & 12.87 & 11.92 & 13.34 &12.42\\
            ZSCL~\cite{zheng2023preventing} & 3.59 & 4.71 & 4.17 & 2.81 & 3.55 & 1.97 & \textbf{1.47} & 2.30 &3.07 \\
            Ours & \textbf{1.44} & \textbf{1.80} & \textbf{1.01} & \textbf{1.53} & \textbf{2.17} & \textbf{1.80} & {1.65} & \textbf{1.82} &\textbf{1.65}\\

            \midrule

            \textbf{Average accuracy ($\uparrow$)} & & & & & & & & \\

            Continual FT & 75.17 & 75.13 & 76.01 & 67.17 & 75.54 & 74.47 & 71.66 & 76.40 &73.94\\
            LwF~\cite{li2017learning} & 74.14 & 77.44 & 77.94 & 74.89 & 77.30 & 77.43 & 75.50 & 76.51 &76.39\\
            iCaRL~\cite{rebuffi2017icarl} & 76.97 & 78.82 & 78.57 & 75.43 & 78.08 & 78.10 & 75.70 & 77.52 &77.40\\
            ZSCL~\cite{zheng2023preventing} & 80.49 & 82.54 & 82.99 & 82.08 & 82.17 & 80.99 & 80.30 & 80.09 &81.46\\
            Ours & \textbf{83.35} & \textbf{83.57} & \textbf{83.88} & \textbf{83.70} & \textbf{84.46} & \textbf{83.82} & \textbf{83.89} & \textbf{83.43} &\textbf{83.76}\\
            
            \bottomrule
            
            \toprule
        \end{tabular}
        }
    \end{adjustbox}
    
\end{table}

\subsubsection{Multi-Domain Task-Incremental Learning.}
We evaluate the efficacy of our proposed method on the Multi-Domain Task-Incremental Learning (MTIL) benchmark~\cite{zheng2023preventing}. This benchmark introduces a sequence of tasks with varying data distributions and distinct label spaces. Following the standard setting of task-incremental learning~\cite{li2019learn}, we evaluate the model on each dataset in a sequential manner during inference.


As shown in \cref{tab:mtil}, we present the quantitative comparisons with different methods on the MTIL benchmark~\cite{zheng2023preventing}. The results demonstrate that our method outperforms SOTA CL approaches, showing the effectiveness of our proposed method. By leveraging dual teachers with the proposed teacher selective mechanism, our framework is able to alleviate catastrophic forgetting on all training sequences with less than $2\%$ of performance degradation and properly preserves the zero-shot classification capability.


\subsubsection{Multi-Domain Class-Incremental Learning.}
As mentioned in ~\cite{wang2022learning, wang2022dualprompt}, the above \emph{task-incremental} learning setting requires the information of task identity (\eg, label space) of each test image, so that might not reflect practical scenarios.
In this work, we further consider a more challenging scenario, Multi-Domain Class-Incremental Learning (MCIL), where the task (data domain) to be evaluated is not known during inference.
To realize this, we conduct a unified label space by merging label spaces from all datasets at the inference stage.

As we can observe in \cref{tab:mcil}, while there is a slight performance drop for all methods, our method consistently surpasses the other SOTA approaches, with about $1\sim 3\%$ improvement for almost all metrics across different sequences. From the above results, we successfully confirm the effectiveness and robustness of our proposed method in the more challenging class-incremental setting.

\subsection{Analysis}
\label{subsec:analysis}

\begin{figure}[t]
\centering
\begin{minipage}[b]{0.47\textwidth}
  \centering
  \includegraphics[width=\textwidth]{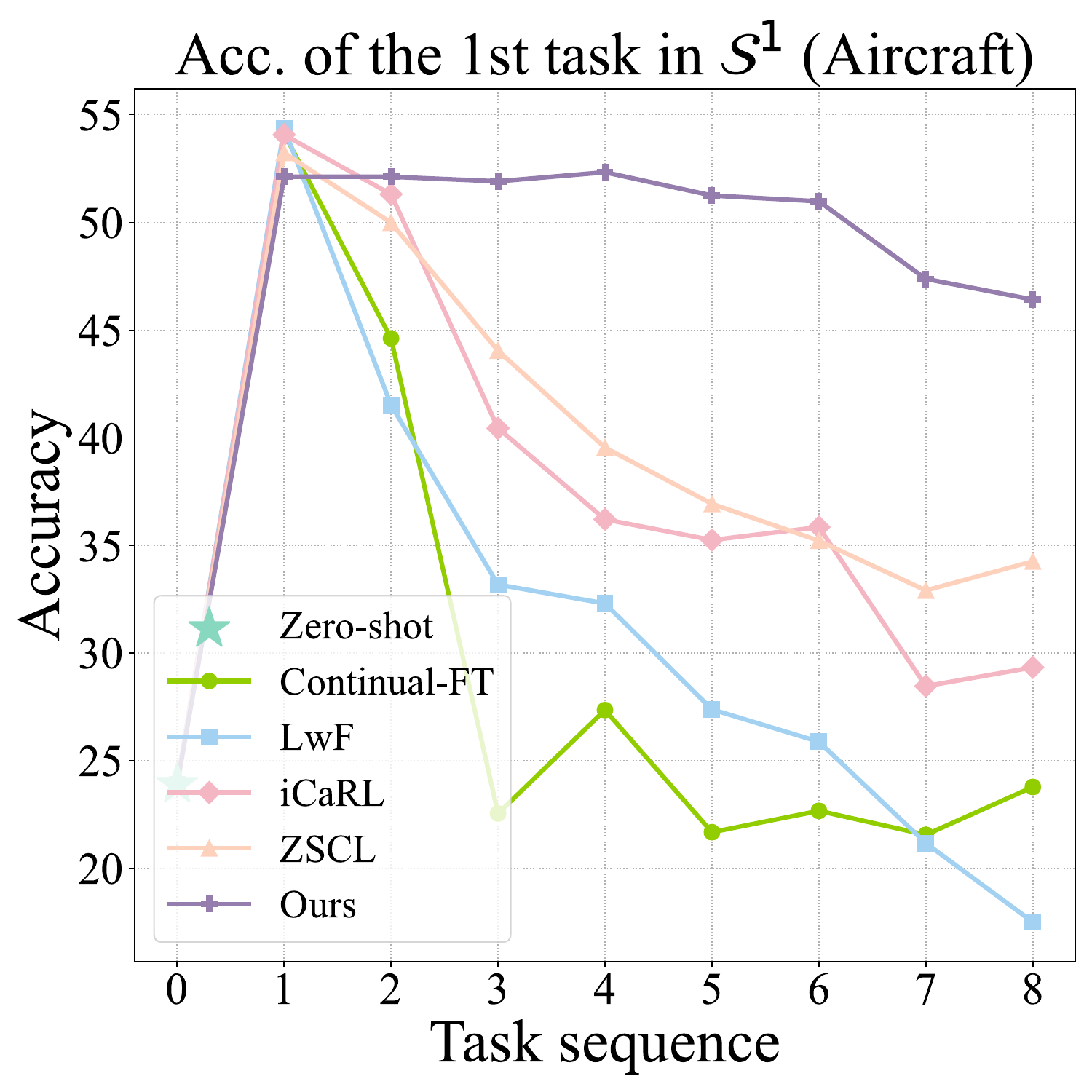}
\end{minipage}
\hspace{0.04\textwidth}
\begin{minipage}[b]{0.47\textwidth}
  \centering
  \includegraphics[width=\textwidth]{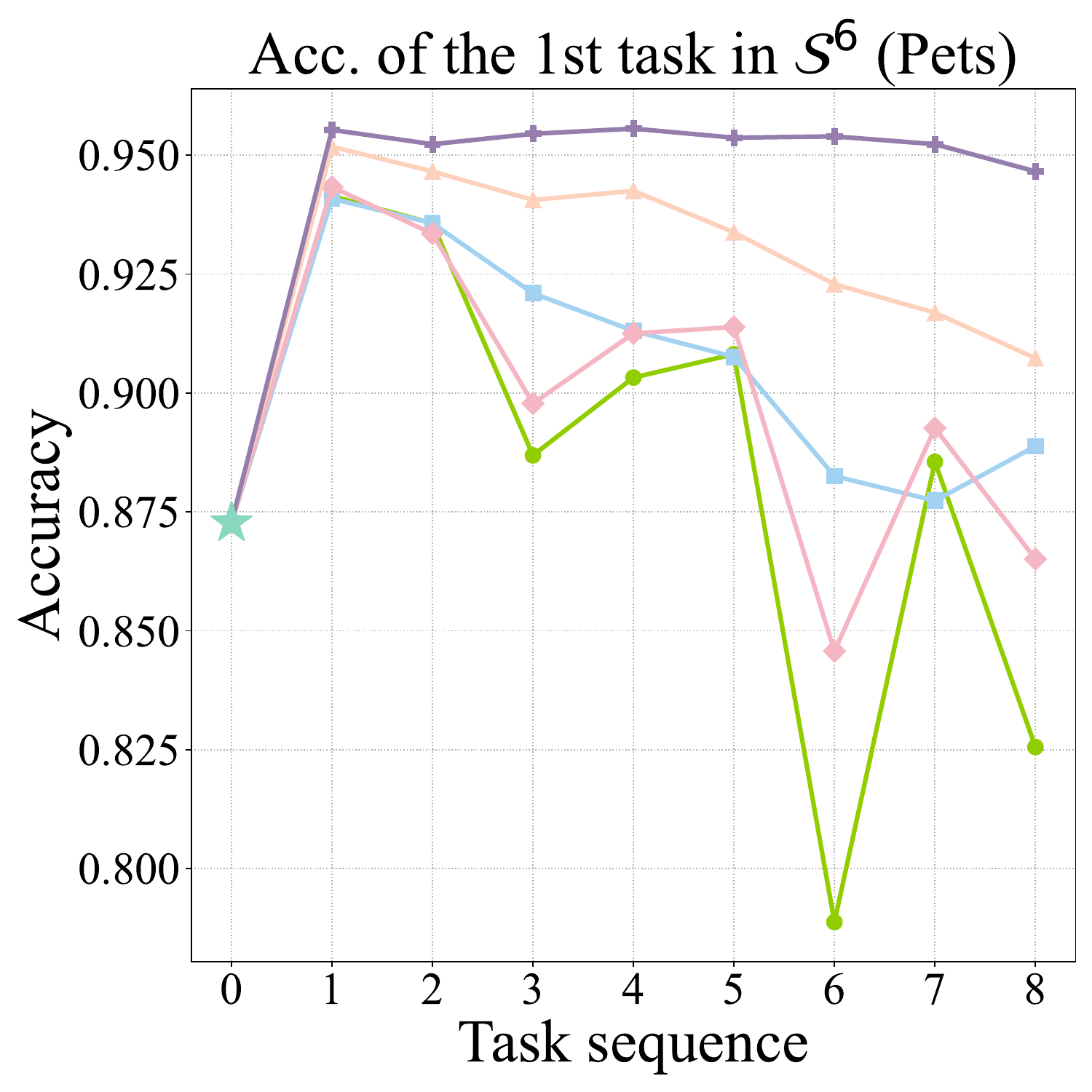}
\end{minipage}
\vspace{-2mm}
\caption{Assessment of catastrophic forgetting with Aircraft (left) and Pets (right) as the first task in the continual learning sequence (i.e., the horizontal axis). It can be seen that our method is able to maintain their accuracies at the end of learning sequence.}

\label{fig:forgetting}
\end{figure}

\subsubsection{Assessment of Catastrophic Forgetting on the first Dataset.}
The performance on the first dataset in a training sequence inevitably suffers from the most severe catastrophic forgetting during continual learning. In \cref{fig:forgetting}, we plot the evaluation results on the first task through all training rounds to visualize the degree of catastrophic forgetting. The results clearly demonstrate that our method effectively maintains stable performance on the initial task even after multiple training rounds, verifying the effectiveness of our method in alleviating catastrophic forgetting on previous tasks. More experimental results are provided in the supplementary.

\begin{figure}[t]
\centering
\begin{minipage}[b]{0.47\textwidth}
  \centering
  \includegraphics[width=\textwidth]{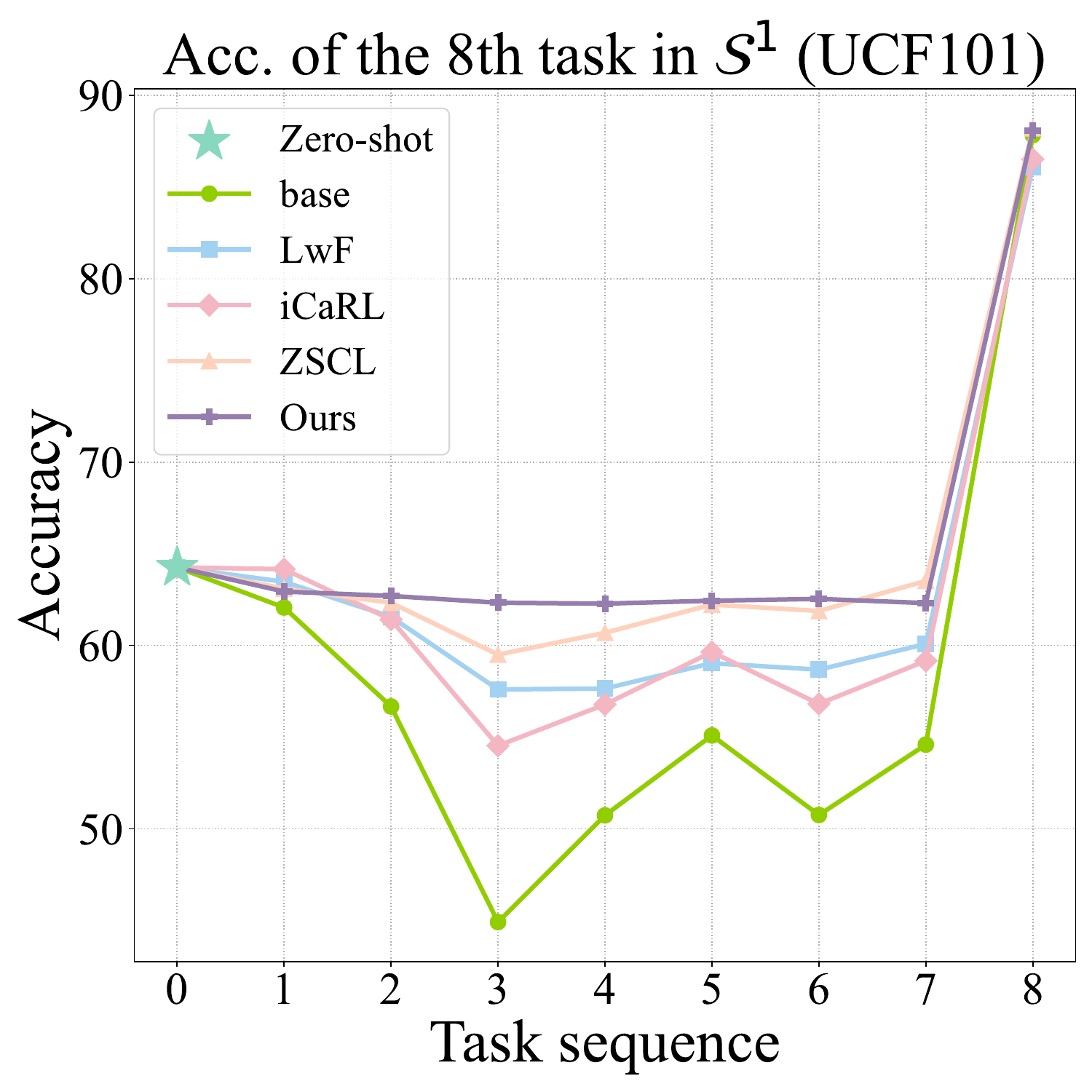}
\end{minipage}
\hspace{0.04\textwidth}
\begin{minipage}[b]{0.47\textwidth}
  \centering
  \includegraphics[width=\textwidth]{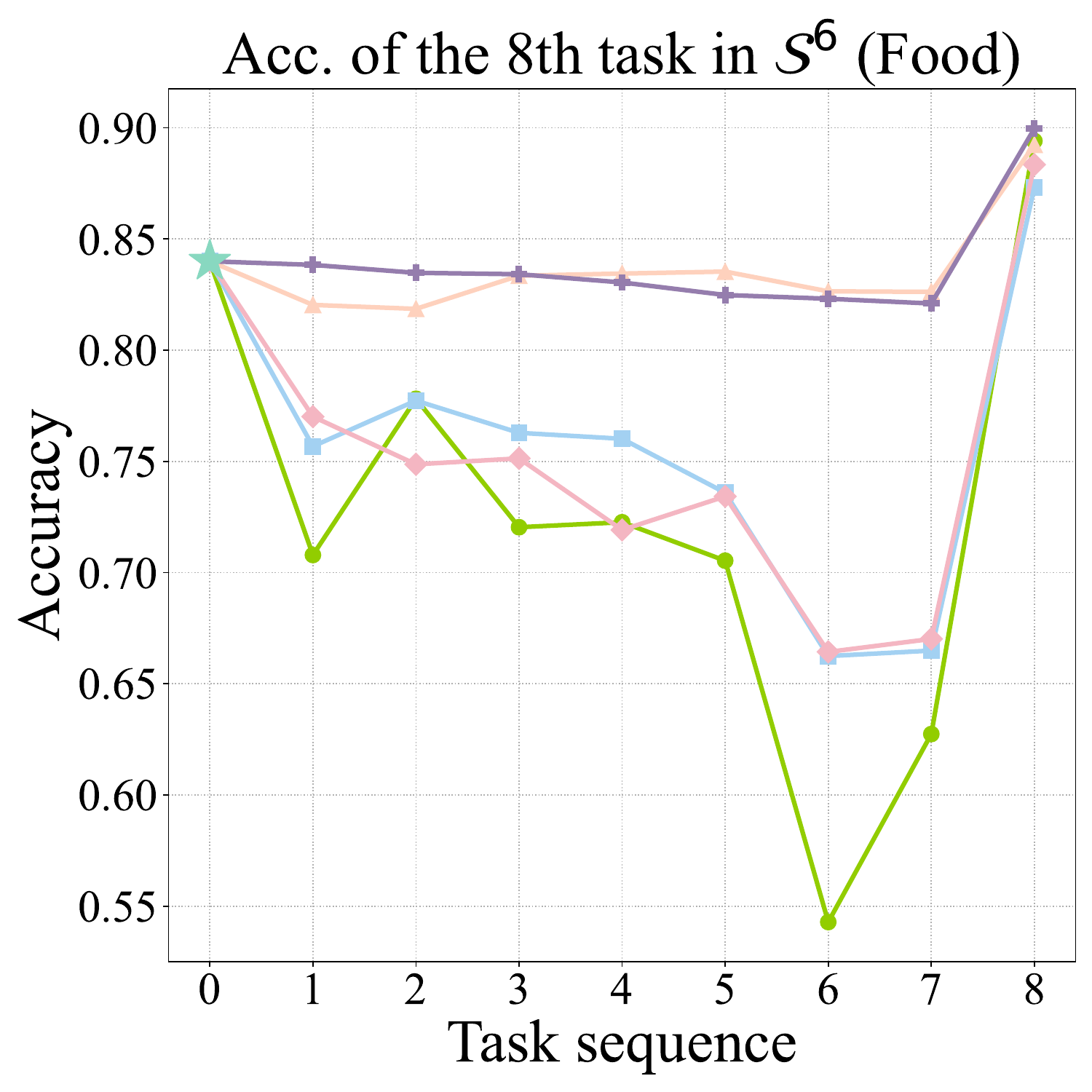}
\end{minipage}
\vspace{-2mm}
\caption{Assessment of zero-shot degradation with UCF101 (left) and Food (right) as the last task in the continual learning sequence (i.e., the horizontal axis). It can be seen that our method shows satisfactory accuracies before finetuning on the last task.}
\label{fig:degradation}
\end{figure}

\subsubsection{Assessment of Zero-Shot Degradation on the last Dataset.}

In addition to maintaining the knowledge learned from previous tasks, it is also crucial to preserve the zero-shot transferability for continual learning on VLMs. 
\cref{fig:degradation} indicates the level of zero-shot degradation for the last dataset, which experiences the most significant zero-shot degradation. The results show that we are able to keep almost the same zero-shot performance compared with the original pre-trained CLIP, demonstrating the effectiveness of our method in preserving the pre-trained zero-shot classification capability during sequential fine-tuning. More examples on different training sequences are shown in the supplementary.

\subsubsection{Empirical Average Dual-Teacher Discrepancy.}
In \cref{subsec:main_method}, we argue that the dual-teacher discrepancy is highly related to determining whether an image is visually similar to previously fine-tuned data. Here, we verify the idea with empirical analysis. At stage $k=2$, given a model $g_1$ fine-tuned on the Aircraft dataset ($\calT^1$) and a pre-trained model $g_0$, we calculate the average dual-teacher discrepancy between $g_1$ and $g_0$ across various datasets, as shown in \cref{tab:avg_dist}.
The results show that the discrepancy $d$ on the previously fine-tuned data (Aircraft, $\calT^1$) is significantly greater than on other datasets not fine-tuned by $g_1$, while data not aligned with $\calX^1$ shows a lower discrepancy, empirically supporting our intuition in \cref{eq:expected_comparison}.

\begin{table}[t]
    \caption{Empirical average dual-teacher discrepancy $d$ between the model $g_{1}$ trained on Aircraft ($\calT^1$) and the orig1inal pre-trained $g_0$. Take Food for example, its discrepancy is calculated by $d(g_0(\xx), g_1(\xx))$ where $\xx$ denotes images from Food. Since $g_{1}$ is finetuned on Aircraft only, a large discrepency score $d$ of 1.059 for Aircraft is expected.}
    \label{tab:avg_dist}
    \begin{adjustbox}{width=\textwidth,center}
    
    \centering
        {\setlength{\tabcolsep}{8pt}
        \begin{tabular}{l|c|ccccccc}
            \bottomrule
            
            \toprule
            Dataset & Aircraft & DTD & EuroSAT & Flowers & Food &Pets & Cars & UCF101 \\
            \midrule
            Distance & 1.059 & 0.090 &0.126 &0.073 &0.091 &0.067 &0.170 &0.112\\

            \bottomrule

            \toprule
        \end{tabular}
        }
    \end{adjustbox}
    
\end{table} 

\begin{table}[t]
    \caption{An ablation study on knowledge distillation from different teacher selections in $\calS^1$. Distilling from either $g_0$ or $g_{k-1}$ leads to unsatisfactory continual learning and zero-shot performance, respectively. Our method, distilling from both teachers, preserves the zero-shot capability and mitigates catastrophic forgetting in previous tasks.}
    \label{tab:ablation}
    \begin{adjustbox}{width=\linewidth,center}
    
    \centering
    {\setlength{\tabcolsep}{15pt}
        \begin{tabular}{lccc}
        \bottomrule
        
        \toprule
        Method & Forgetting ($\downarrow$)   & Degradation ($\downarrow$)  & Avg. Accuracy ($\uparrow$) \\

        \midrule
        Distill from $g_0$ & 5.26 & 2.51 & 81.35 \\
        Distill from $g_{k-1}$ & 2.63 & 3.36 & 83.61\\

        Ours &\textbf{1.70} &\textbf{1.55} &\textbf{84.48} \\

        \bottomrule

        \toprule
        \end{tabular}}
\end{adjustbox}
\end{table}

\subsubsection{Ablation study to the choice of different teachers.}

To verify the performance improvement of our proposed dual-teacher distillation mechanism, we conduct an ablation study to different choices of teacher models and present the results in \cref{tab:ablation}. It can be seen that while distilling from the pre-trained model ($g_0$) results in satisfactory zero-shot performance with a drop of $2.51$, it fails to prevent catastrophic forgetting. Conversely, distilling from the most recent model $g_{k-1}$ preserve the continual learning performance but compromise zero-shot capability. Our method, an adaptive distillation scheme, is shown to alleviate catastrophic forgetting while preserving zero-shot performance.

\subsubsection{Visualization of Reference Images with Large Selection Scores $\eta$.}

\begin{figure}[t]
\centering
\begin{minipage}{0.4\textwidth}
  \centering
  \includegraphics[width=\textwidth]{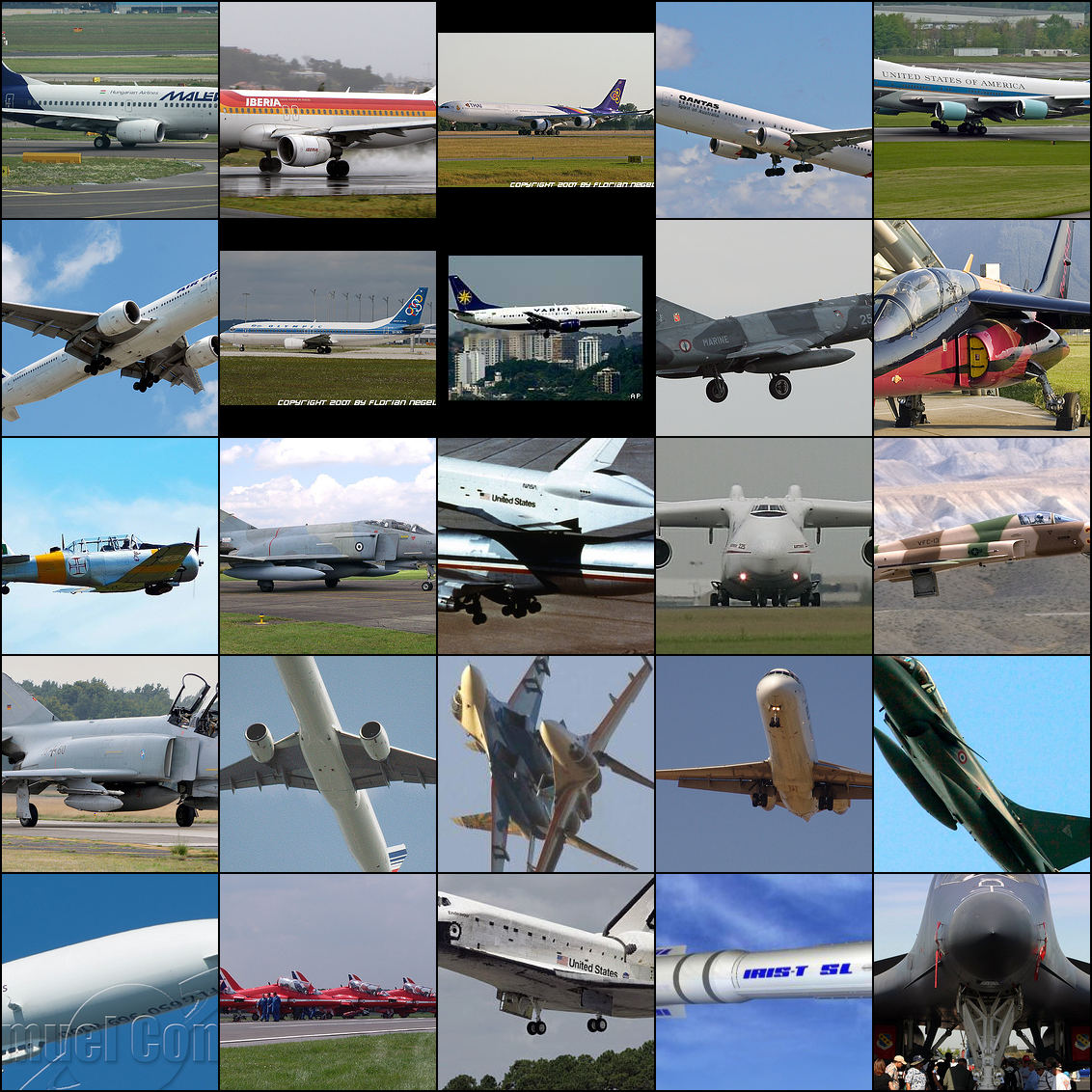}
\end{minipage}
\begin{minipage}{0.12\textwidth}
  \centering
  $\boldsymbol{\rightarrow \cdots \rightarrow}$
\end{minipage}
\begin{minipage}{0.4\textwidth}
  \centering
  \includegraphics[width=\textwidth]{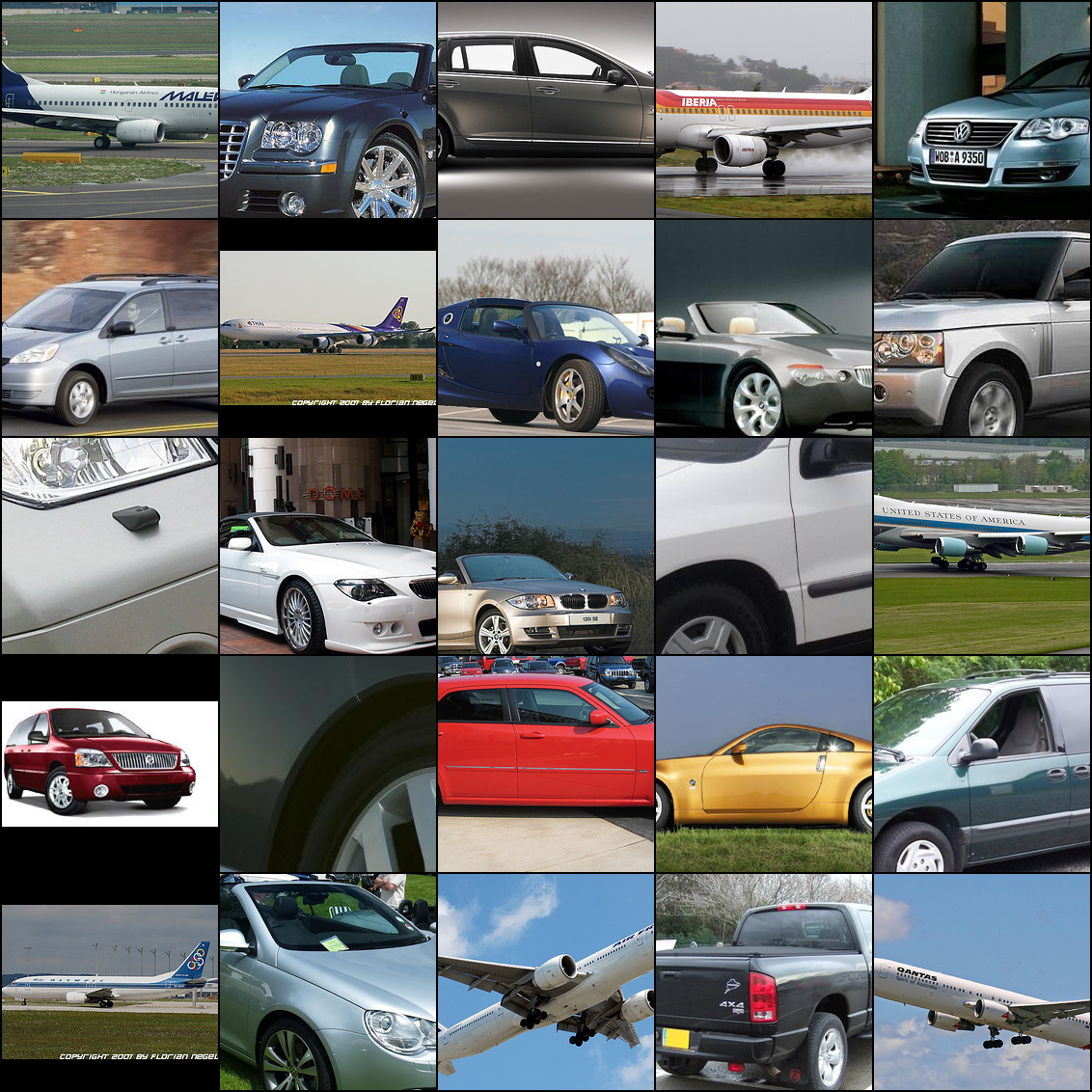}
\end{minipage}

\caption{Example images selected from the reference dataset with large $\eta$ scores. \textbf{Left}: Top-25 reference images selected \textit{after} fine-tuning on the 1st task of Aircraft. This suggests how we utilize such data to prevent possible catastrophic forgetting of Aircraft. \textbf{Right}: Top-25 reference images selected after continual learning across all tasks (with the last task as Cars). It can be seen that our scheme still selects reference images closer to prior tasks (e.g., Aircraft), explaining how zero-shot transfer ability is preserved.}

\label{fig:visual}
\end{figure}

Our proposed selection function $\eta$ (\cref{eq:similarity_estimation}) aims to select the appropriate teacher by estimating the visual similarity between a reference image and previously fine-tuned data of $g_{k-1}$. We verify the effectiveness of the selection function by selecting Top-K reference images with large $\eta$ scores. \cref{fig:visual} visualize Top-25 reference images selected after training on certain datasets. We highlight two observations from the visualized results in the following,

\begin{itemize}
    \item As shown in the left half of \cref{fig:visual}, after fine-tuning the model $g_1$ on the 1st task (\ie, Aircraft), the Top-25 reference images with large $\eta$ scores are highly similar to previously fine-tuned task without actually accessing any information related to previous data.

    \item After the model has been sequentially trained on multiple rounds (with the last task as Cars), our scheme is still able to select reference images closer to prior tasks (\eg, Aircraft), demonstrating the ability to preserve the earliest fine-tuned knowledge even after multiple downstream tasks, as visualized in the right part of \cref{fig:visual}.
\end{itemize}


\section{Conclusion}
\label{sec:conclusion}
In this paper, we propose a Selective Dual-Teacher Knowledge Transfer framework for continual learning that tackles catastrophic forgetting and preserves zero-shot generalization ability simultaneously. By leveraging the most recent fine-tuned and original pre-trained VLMs as dual teachers, our framework selectively distills knowledge based on a dual-teacher discrepancy observed from an auxiliary reference dataset, without requiring label supervision. Comprehensive experiments, including comparisons with state-of-the-art continual learning methods and extensive analysis, quantitatively and qualitatively verify the effectiveness of our framework over existing approaches.

\subsubsection{Limitation.}

Following~\cite{zheng2023preventing}, our work leverages an unlabeled reference dataset with the proposed teacher selection mechanism to identify the proper teacher network for knowledge distillation. If the reference dataset is very different from the previously fine-tuned tasks (e.g., medical images), the most recent fine-tuned VLM ($g_{k-1}$) would rarely be selected during training, so that the catastrophic forgetting might not be addressed well. 

To further assess the impact of different reference datasets, we leave additional experimental results in the supplementary material due to page limitations. For example, subsets of larger-scale datasets such as ConceptualCaptioning 12M~\cite{changpinyo36conceptual} and LAION 5B~\cite{schuhmann2022laion}) can be further considered and exploited. 

\subsubsection{Acknowledgement.}

This work is supported in part by the National Science and Technology Council via grant NSTC 112-2634-F-002-007, NSTC 113-2640-E-002-003 and the Featured Area Research Center Program within the framework of the Higher Education Sprout Project by the Ministry of Education 113L900902. We also thank the National Center for High-performance Computing (NCHC) for providing computational and storage resources.

%
%
\bibliographystyle{Styles/splncs04}
\bibliography{Styles/egbib}

\clearpage
\noindent\textbf{\large Appendix}
\appendix

\section{Evaluation Details}

\subsubsection{Datasets Statistics.}

We provide the detailed statistics of 8 fine-grained datasets and the reference dataset (\ie, ImageNet~\cite{deng2009imagenet}) in \cref{tab:datasets}. The splits for training, validation and test of each dataset basically follow the setting provided by Zhou \etal~\cite{zhou2022learning}. Following the setting proposed in ZSCL~\cite{zheng2023preventing}, we sample 100,000 unlabeled images from ImageNet as the reference dataset.

\subsubsection{Details of Multiple Training Sequences.} We introduce \emph{Multip Training Sequences} evaluation protocol to thoroughly evaluate every method over different training sequences in \cref{subsec:evaluation}. Here we provide the detailed order of tasks for each sequence in \cref{tab:sequences}.

\section{More Implementation Details}
\subsubsection{Re-Weighted Dual-Teacher Knowledge Distillation Loss.}
Our proposed Dual-Teacher Knowledge Distillation loss shows the way to select the appropriate teacher model for a reference image according to the dual-teacher discrepancy and selection score $\eta$. In practice, there are few reference images with higher dual-teacher discrepancy. To address this potential imbalance problem, we apply a loss re-weighting strategy~\cite{cui2019class} as a post-processing technique. Specifically, the re-weighted dual-teacher knowledge distillation loss is shown below:
\begin{equation}
    \tilde{\calL}^{\text{dual}}_{\text{KD}}  = \lambda \cdot \sum_{\xx \sim \calX^{\text{ref}}} \eta(\xx) \cdot \calL^{k-1}_{\text{KD}} + \sum_{\xx \sim \calX^{\text{ref}}} (1-\eta(\xx)) \cdot \calL^0_{\text{KD}},
\end{equation}
where $\lambda$ is a hyper-parameter to control the imbalance ratio between the KD loss to the most recent fine-tuned model $g_{k-1}$ and the KD loss to the pre-trained model $g_0$. Emprically we set $\lambda = 9$ to properly deal with the imbalance issue for every experiment in this work.

\subsubsection{Hyper-Parameters to the $\eta$ Selection Function.}

\begin{table}[t]
    \caption{Detailed statistics for each dataset.}
    \label{tab:datasets}
    \begin{adjustbox}{width=\textwidth,center}
    
    \centering
        {\setlength{\tabcolsep}{20pt}
        \begin{tabular}{lcccccccc}
            \bottomrule
            
            \toprule
            Dataset & Classes & Train & Val & Test \\
            \midrule

            ImageNet~\cite{deng2009imagenet} &1,000 &1.28M & N/A &50,000 \\
            \midrule
            \midrule
            Aircraft~\cite{maji2013fine} & 100 & 3,334 &3,333 &3,333 \\
            DTD~\cite{cimpoi2014describing} &47 &2,820 &1,128 &1,692 \\
            EuroSAT~\cite{helber2019eurosat} &10 &13,500 &5,400 &8,100 \\
            Flowers-102~\cite{nilsback2008automated} &102 &4,093 &1,633  &2,463 \\
            Food-101~\cite{bossard2014food} &101 &50,500 &20,200 &30,300 \\
            Oxford-Pets~\cite{parkhi2012cats} &37 &2,944 &736 &3,669 \\
            Stanford-Cars~\cite{krause20133d} &196 &6,509 &1,635 &8,041 \\
            UCF-101~\cite{soomro2012ucf101} &101 &7,639 &1,898 &3,783 \\

            \bottomrule

            \toprule
        \end{tabular}
        }
    \end{adjustbox}
    
\end{table}

\begin{table}[htbp]
\centering
\caption{The order of tasks for each training sequence.}

\begin{adjustbox}{width=\textwidth,center}
{\setlength{\tabcolsep}{5pt}
\begin{tabular}{cllllllll}

\bottomrule

\toprule
Sequence & 1st Task &2nd Task &3rd Task &4th Task &5th Task &6th Task &7th Task &8th Task \\

\midrule

$\calS^1$ & Aircraft & DTD & EuroSAT & Flowers & Food & Pets & Cars & UCF101 \\
$\calS^2$ & DTD & EuroSAT & Flowers & Food & Pets & Cars & UCF101 & Aircraft \\
$\calS^3$ & EuroSAT & Flowers & Food & Pets & Cars & UCF101 & Aircraft & DTD \\
$\calS^4$ & Flowers & Food & Pets & Cars & UCF101 & Aircraft & DTD & EuroSAT \\
$\calS^5$ & Food & Pets & Cars & UCF101 & Aircraft & DTD & EuroSAT & Flowers \\
$\calS^6$ &Pets & Cars & UCF101 & Aircraft & DTD & EuroSAT & Flowers & Food \\
$\calS^7$ & Cars & UCF101 & Aircraft & DTD & EuroSAT & Flowers & Food & Pets \\
$\calS^8$ & UCF101 & Aircraft & DTD & EuroSAT & Flowers & Food & Pets & Cars \\

\bottomrule

\toprule
\end{tabular}}
\end{adjustbox}

\label{tab:sequences}
\end{table}

\begin{table}[t]
\caption{Sensitivity analysis on $\calS^1$ to the hyper-parameters $\delta$ and $\gamma$ in the $\eta$ selection function. We highlight the results of our default setting across all experiments in the main paper in light red.}
    \label{tab:normalize}
    \begin{adjustbox}{width=\textwidth,center}
    
    \centering
        {\setlength{\tabcolsep}{20pt}
        \begin{tabular}{ccccc}

\bottomrule

\toprule
$\delta$ & $\gamma$ & Forgetting ($\downarrow$)& Degradation ($\downarrow$)& Avg. Accuracy ($\uparrow$)\\
\midrule
 & 1/3 & 1.72 & 1.58 & 84.42 \\
 & 1/6 & 1.68 & 1.57 & 84.43 \\
\multirow{-3}{*}{0.1} & 1/9 & \textbf{1.65} & 1.58 & 84.47 \\
\midrule
 & 1/3 & 1.67 & 1.60 & 84.46 \\
 & 1/6 & \cellcolor{lightred}1.70 & \cellcolor{lightred}\textbf{1.55} & \cellcolor{lightred}\textbf{84.48} \\
\multirow{-3}{*}{0.2} & 1/9 & 1.82 & 1.86 & 84.31 \\
\midrule
 & 1/3 & 1.81 & 1.52 & 84.23 \\
 & 1/6 & 2.13 & 1.99 & 84.03 \\
\multirow{-3}{*}{0.3} & 1/9 & 2.45 & 1.99 & 83.93\\
\bottomrule

\toprule

\end{tabular}}
\end{adjustbox}
\end{table}

\begin{table}[t]
    \caption{The performance of different reference datasets with varying size. The default setting for all experiments is marked in light red.}
    \label{tab:ref_dataset}
    \begin{adjustbox}{width=\textwidth,center}
    
    \centering
        {\setlength{\tabcolsep}{10pt}
        \begin{tabular}{lcccc}
        \bottomrule
        
        \toprule
        Ref. Dataset & Size & Forgetting ($\downarrow$)   & Degradation ($\downarrow$)  & Avg. Accuracy ($\uparrow$) \\

        \midrule
        \multirow{3}{*}{ImageNet} & 10k  & 1.92 & 2.12 & 84.18 \\
                                  & \cellcolor{lightred}100k & \cellcolor{lightred}1.70 & \cellcolor{lightred}1.55 & \cellcolor{lightred}84.48 \\
                                  & 200k & 1.65 & \textbf{1.11} & 84.80 \\
        \midrule
        \multirow{3}{*}{Conceptual Captions 12M} & 10k  & 2.28 & 2.17  & 83.84 \\
                                                 & 100k & \textbf{1.50} & 1.88 & 84.48 \\
                                                 & 200k & 1.60 & 1.25 & \textbf{84.99} \\
        \bottomrule

        \toprule
        \end{tabular}}
\end{adjustbox}
\end{table}

Our proposed $\eta$ selection function:

\begin{equation}
    \eta(\xx) = \sigma(\frac{d(g_{k-1}(\xx), g_0(\xx)) - \delta}{\gamma}),
\end{equation}
involves two hyper-parameters: $\delta$ and $\gamma$. At a high-level, $\delta$ serves as a threshold that determining whether to select more from $g_{k-1}$ or $g_0$. As the threshold $\delta$ increases, more reference data points are likely to be assigned values lower than $0.5$, \ie, select KD Loss more from $g_0$. On the other hand, $\gamma$ works as a scaling factor to scale the value before applying the sigmoid function. As $\gamma \rightarrow 0$, the selection function move towards a \textit{hard selection} mechanism, where the $\eta$ scores tend to output either $1$ or $0$, depending on the discrepancy $d(g_{k-1}(\xx), g_0(\xx))$.

\cref{tab:normalize} provides a sensitivity analysis for hyper-parameters $\delta$ and $\gamma$. In general, the performance shows no significant difference when $\delta = 0.1$ or $0.2$, hinting that it is stable enough for a proper range. By default, we select $\delta =0.2$ and $\gamma = 1/6$ across all experiments in this work.

\begin{figure}[t]
\centering
\begin{minipage}{0.4\textwidth}
  \centering
  \includegraphics[width=\textwidth]{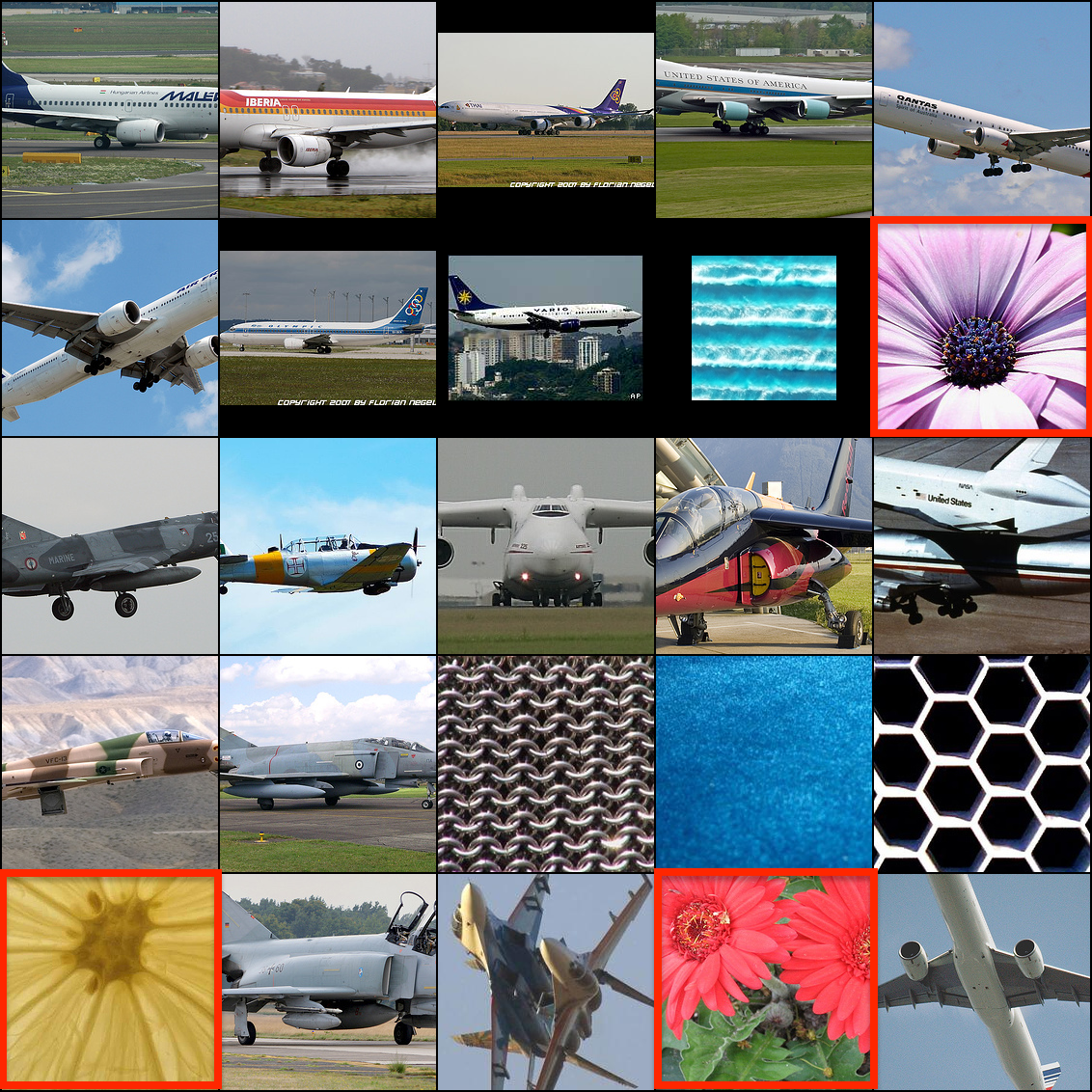}
\end{minipage}
\begin{minipage}{0.1\textwidth}
$\ $
\end{minipage}
\begin{minipage}{0.4\textwidth}
  \centering
  \includegraphics[width=\textwidth]{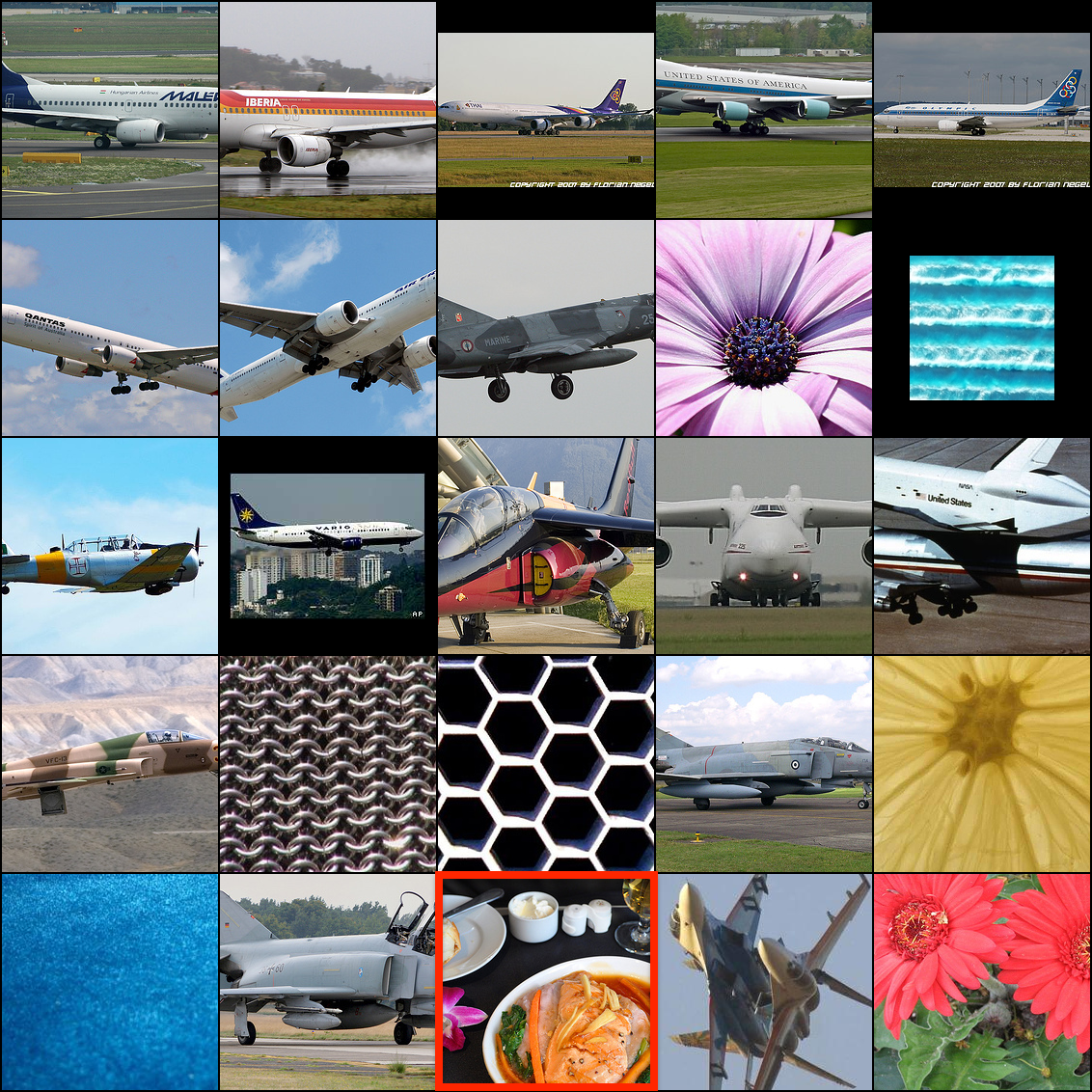}
\end{minipage}

\caption{Example images selected from the reference dataset with large $\eta$ scores. \textbf{Left}: Top-25 reference images selected \textit{after} fine-tuning on the Flowers Dataset. \textbf{Right}: Top-25 reference images selected \textit{after} fine-tuning on the Food Dataset.}

\label{fig:supp_visual}
\end{figure}

\section{Different Choices of Reference Datasets}
Our \emph{Selective Dual-Teacher Knowledge Transfer} framework leverages an unlabeled reference dataset, following the settings in \cite{zheng2023preventing}. As mentioned in the limitation, the composition and the diversity of the images in the reference dataset might greatly affect the final performance. To examine the effect, we conduct ablation studies using different reference datasets (e.g., ConceptualCaptioning 12M~\cite{changpinyo2021conceptual}) and exploring the impact of varying the size of the reference dataset. \cref{tab:ref_dataset} shows the performance of different reference datasets with varying size. While increasing the size of the reference dataset typically enhances performance, empirically there are no significant differences when the size exceeds 100k. By default, we use ImageNet with 100k images as our reference dataset, which also aligns with the same settings in \cite{zheng2023preventing}.

\begin{figure}[t]
\centering
\begin{minipage}[b]{0.47\textwidth}
  \centering
  \includegraphics[width=\textwidth]{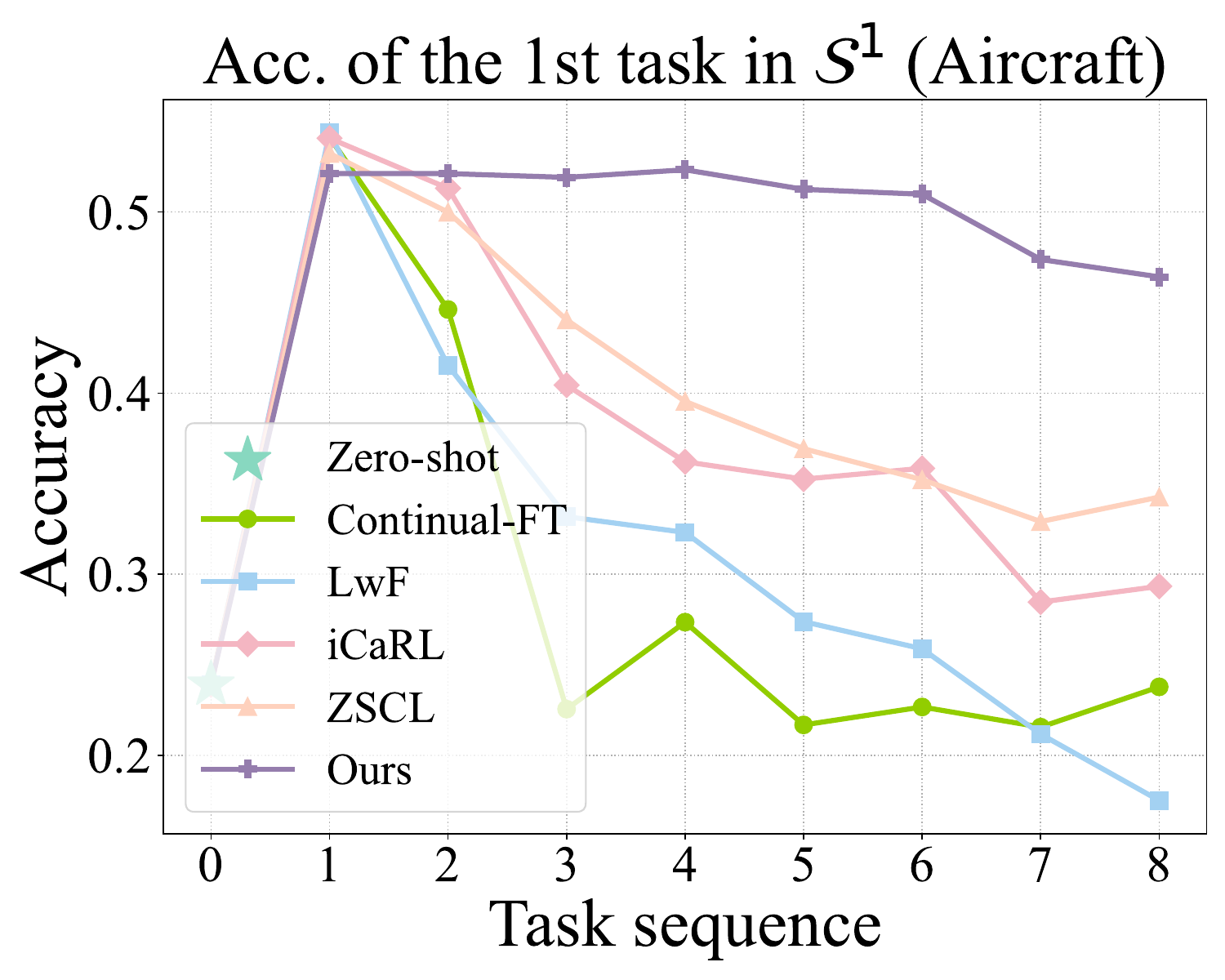}
\end{minipage}
\hspace{0.04\textwidth}
\begin{minipage}[b]{0.47\textwidth}
  \centering
  \includegraphics[width=\textwidth]{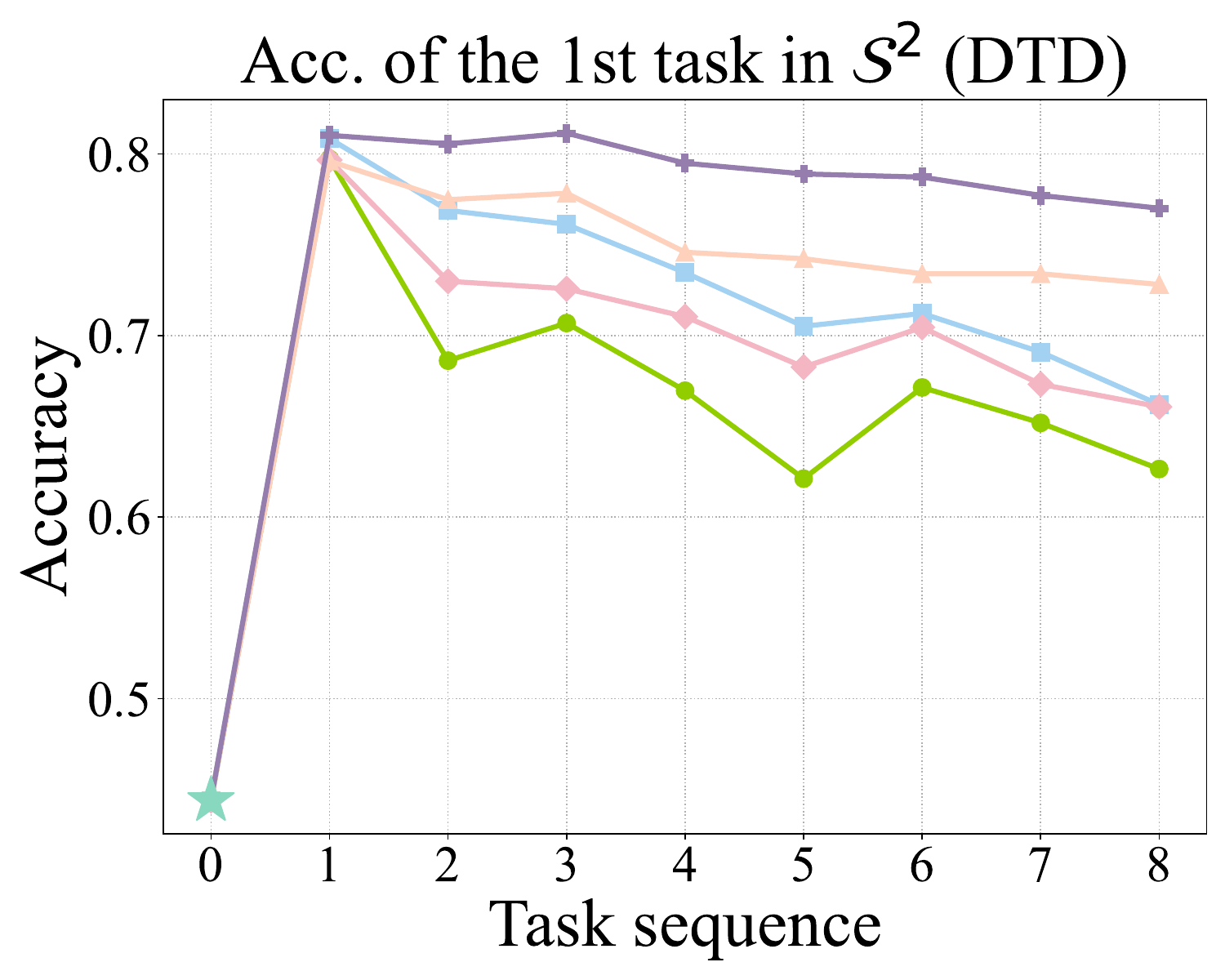}
\end{minipage}

\begin{minipage}[b]{0.47\textwidth}
  \centering
  \includegraphics[width=\textwidth]{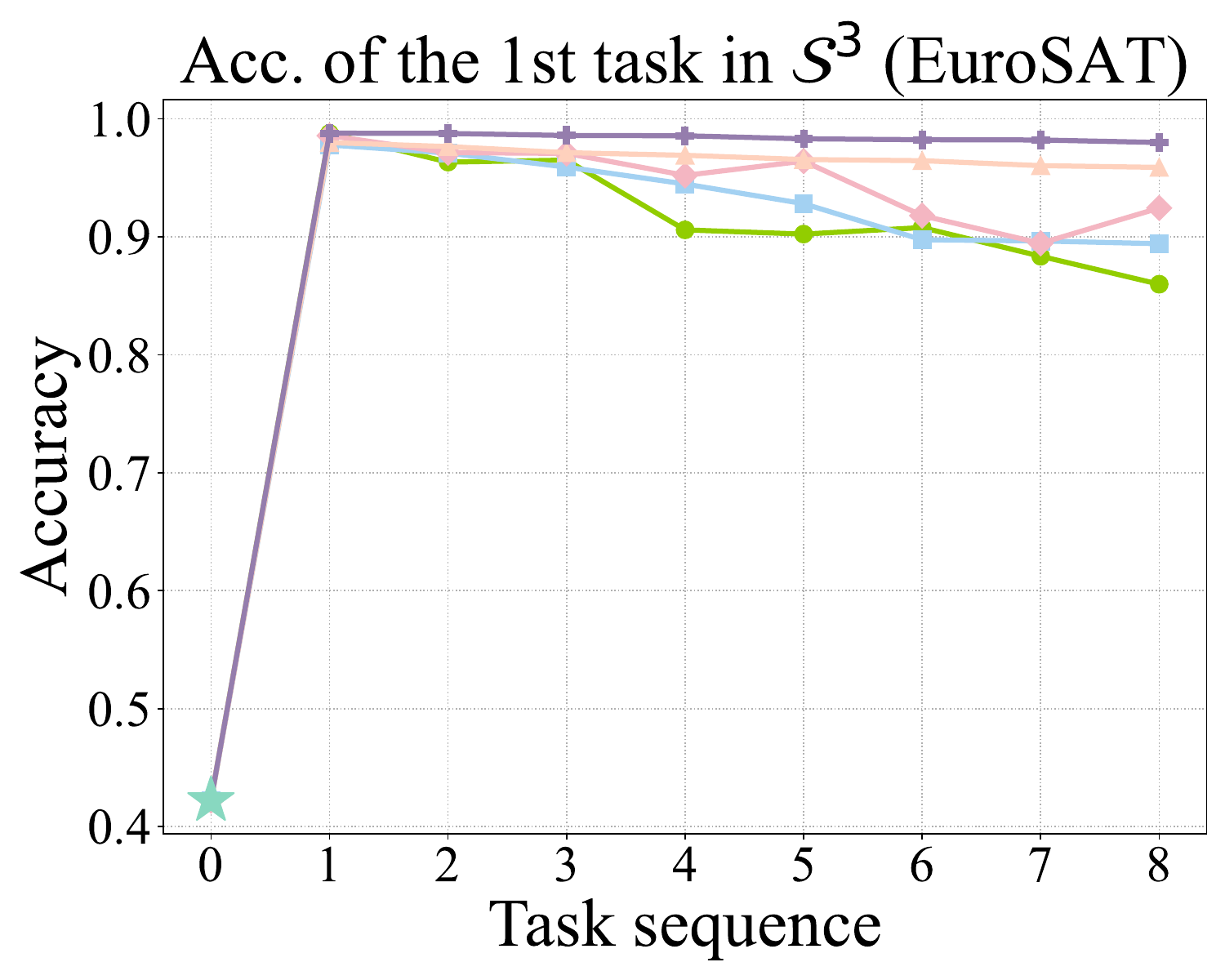}
\end{minipage}
\hspace{0.04\textwidth}
\begin{minipage}[b]{0.47\textwidth}
  \centering
  \includegraphics[width=\textwidth]{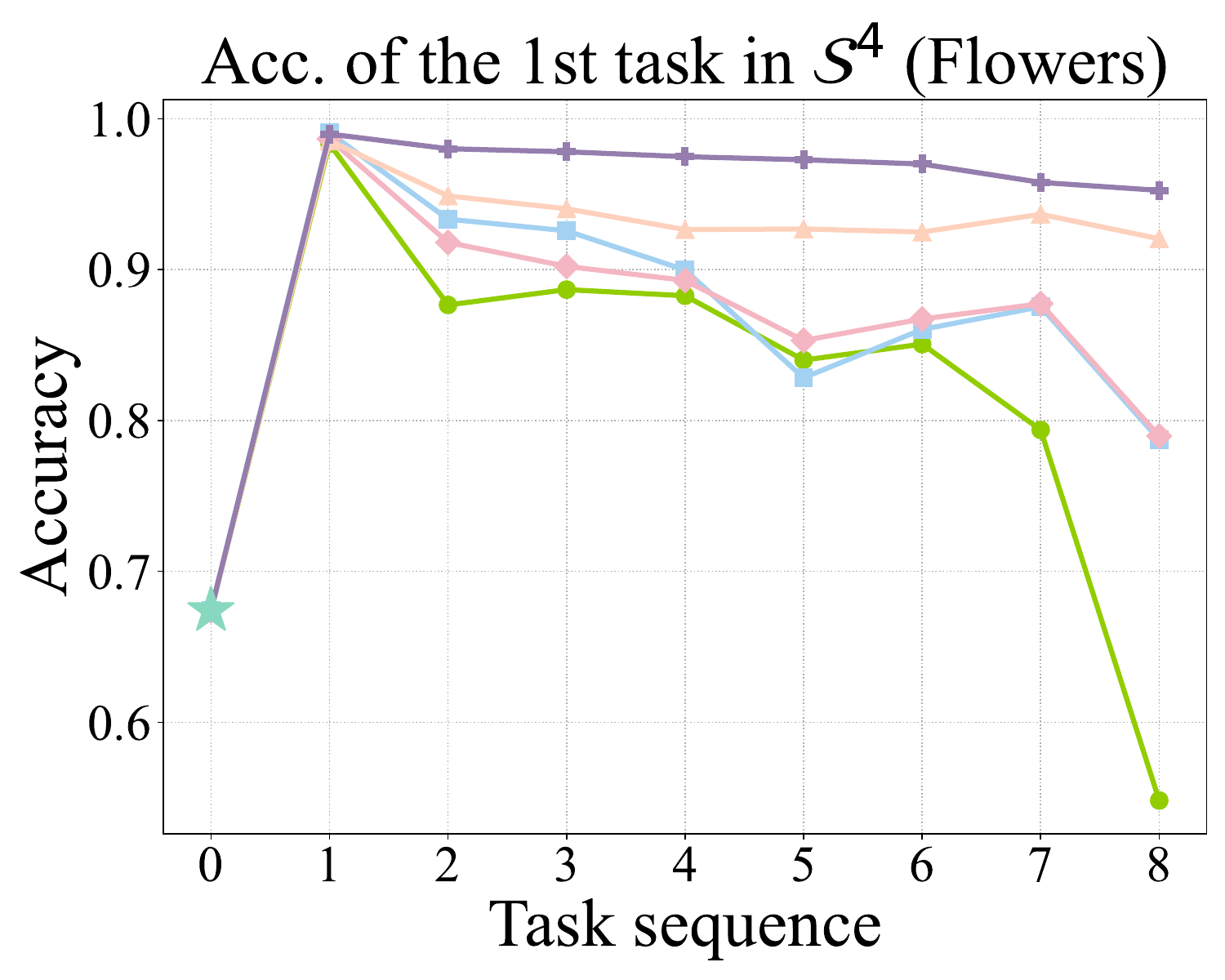}
\end{minipage}

\begin{minipage}[b]{0.47\textwidth}
  \centering
  \includegraphics[width=\textwidth]{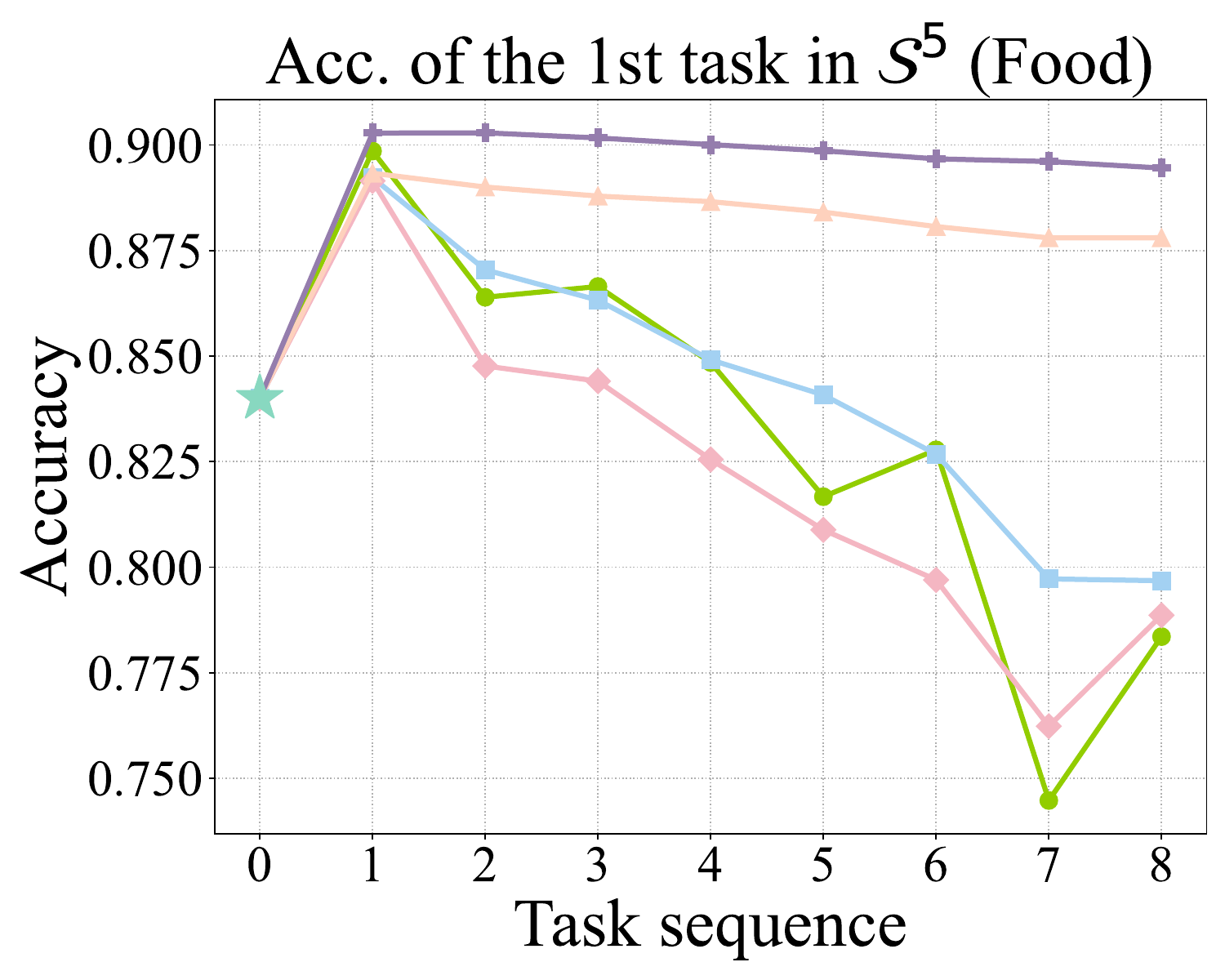}
\end{minipage}
\hspace{0.04\textwidth}
\begin{minipage}[b]{0.47\textwidth}
  \centering
  \includegraphics[width=\textwidth]{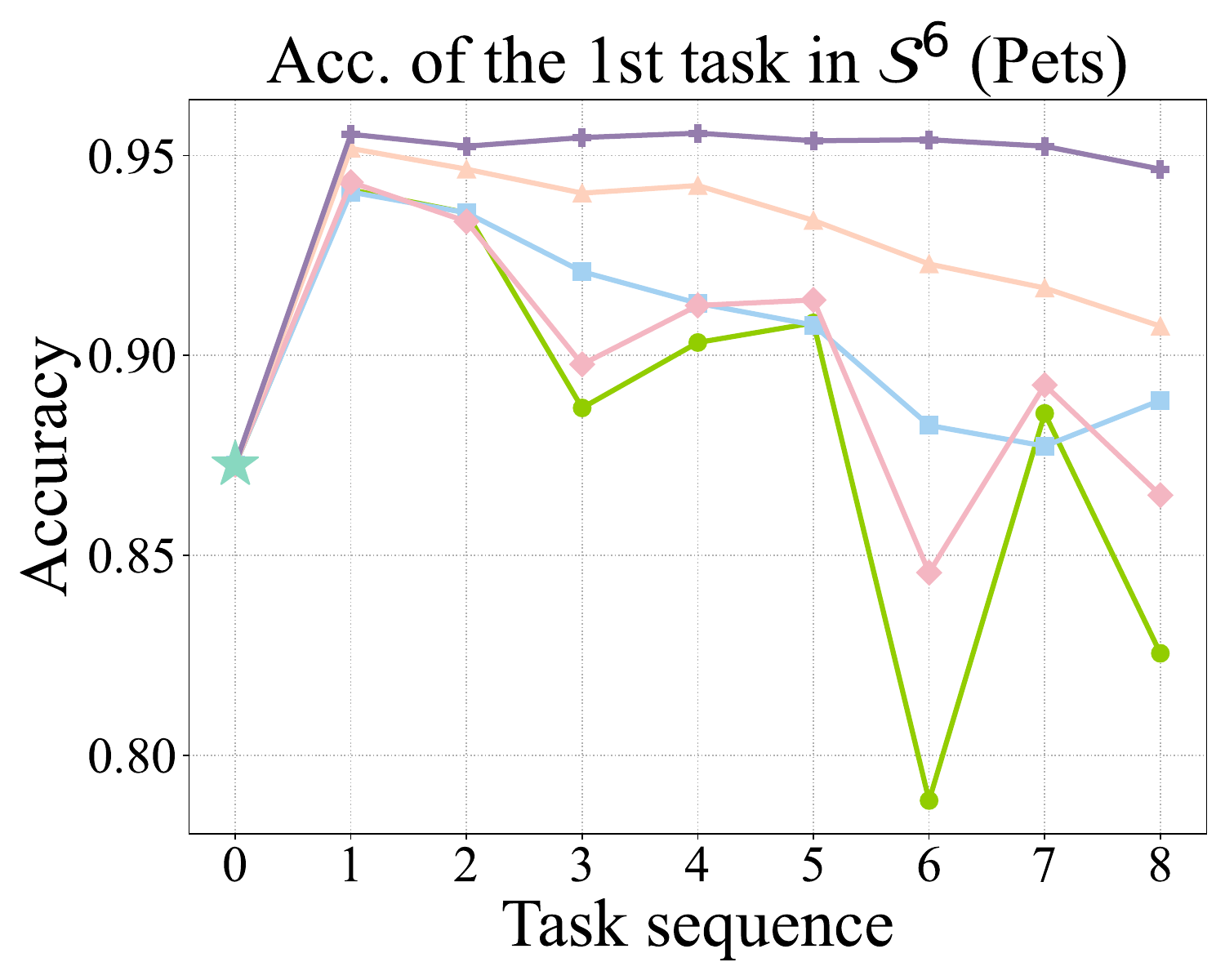}
\end{minipage}

\begin{minipage}[b]{0.47\textwidth}
  \centering
  \includegraphics[width=\textwidth]{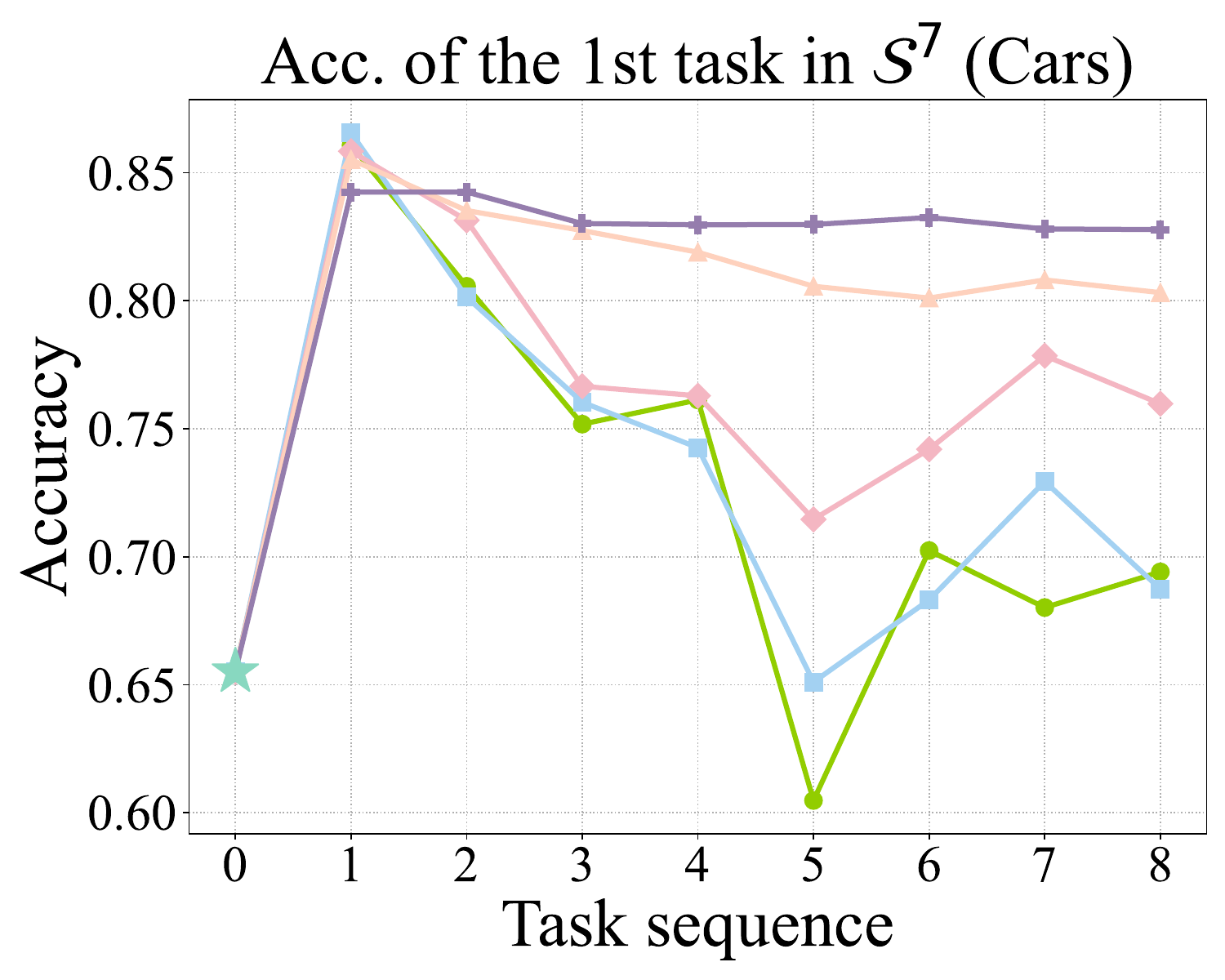}
\end{minipage}
\hspace{0.04\textwidth}
\begin{minipage}[b]{0.47\textwidth}
  \centering
  \includegraphics[width=\textwidth]{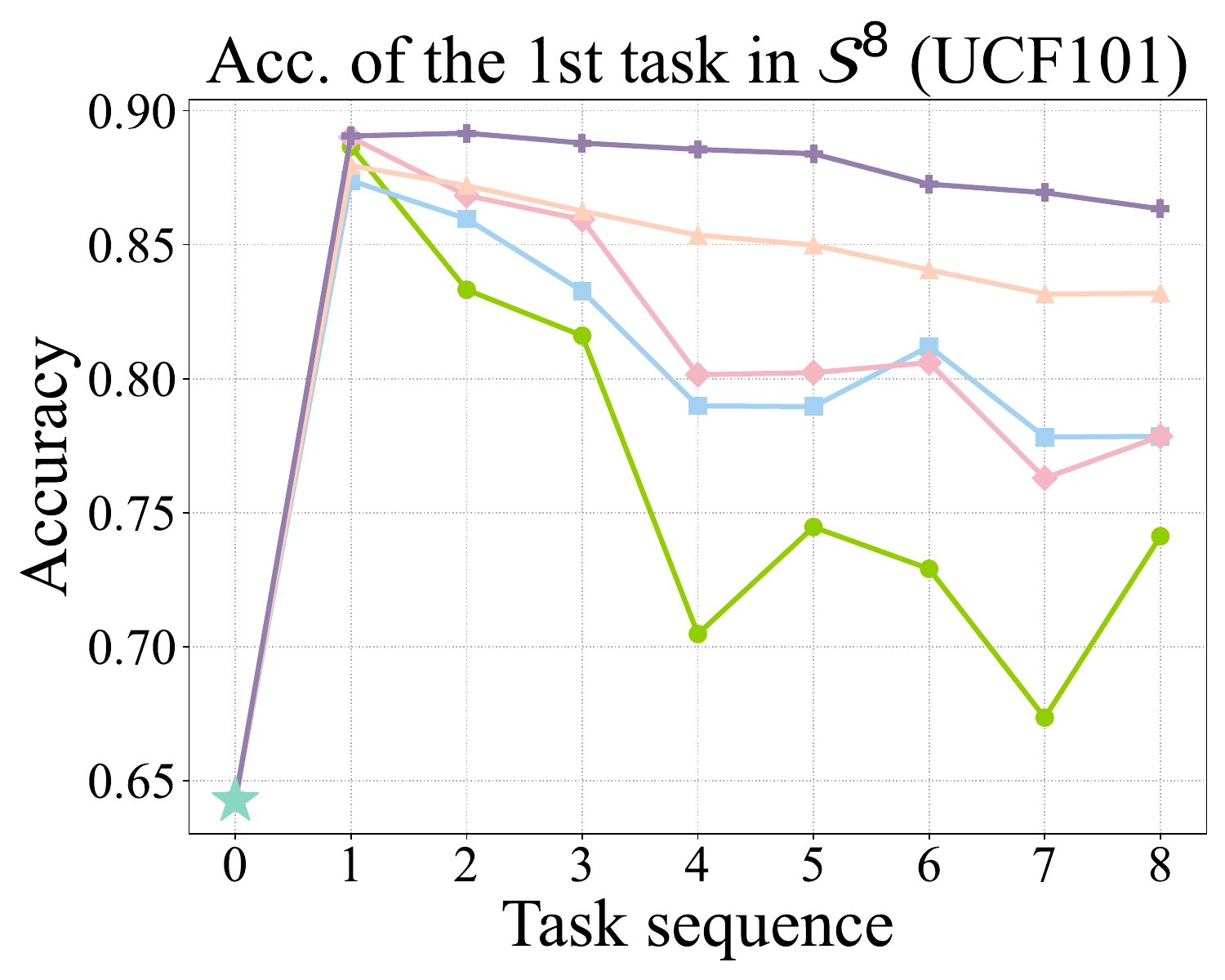}
\end{minipage}

\caption{Assessment of catastrophic forgetting with the first task in the continual learning sequence (i.e., the horizontal axis). It can be seen that our method is able to maintain their accuracies at the end of learning sequence.}

\label{fig:supp_forgetting}
\end{figure}

\begin{figure}[t]
\centering
\begin{minipage}[b]{0.47\textwidth}
  \centering
  \includegraphics[width=\textwidth]{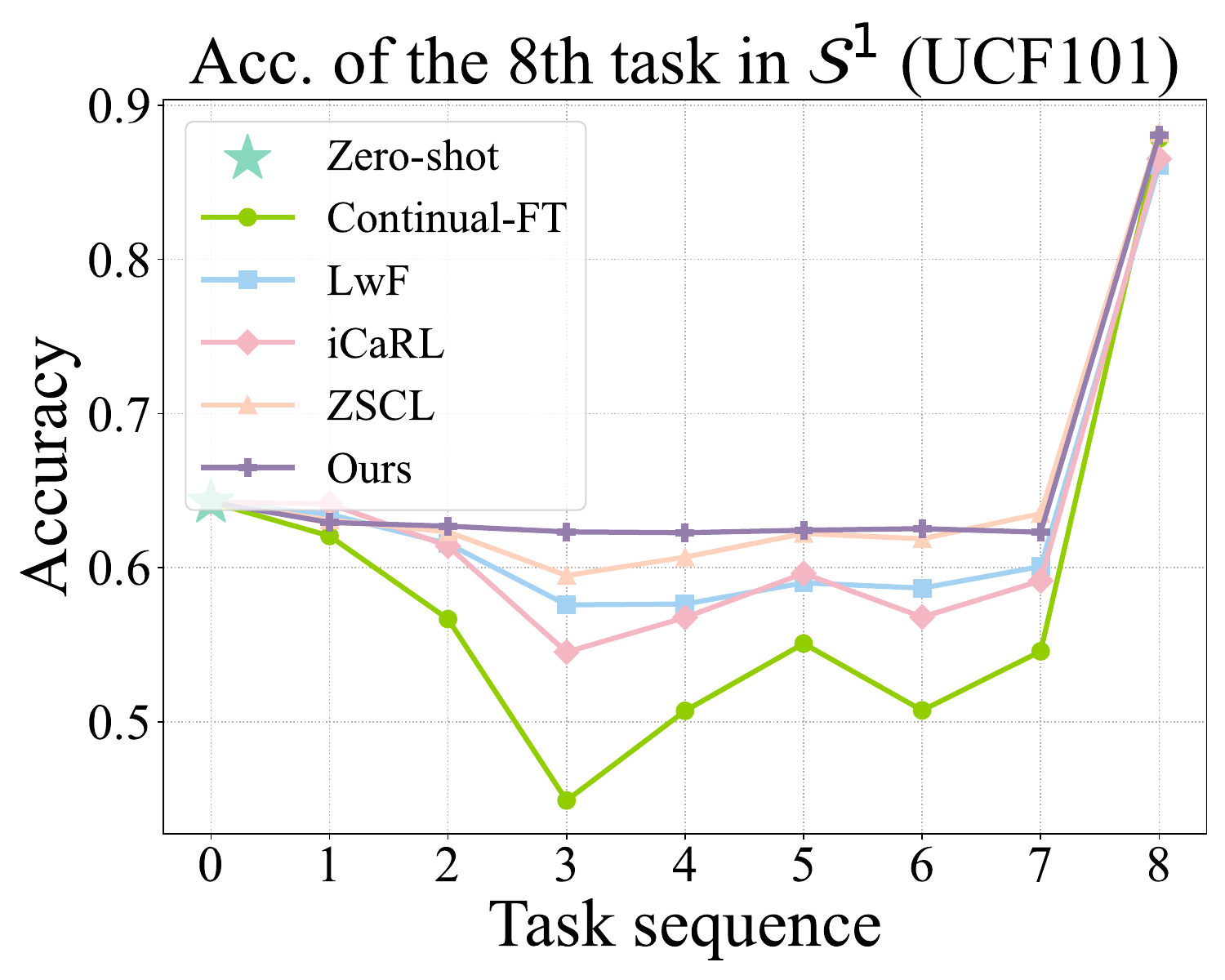}
\end{minipage}
\hspace{0.04\textwidth}
\begin{minipage}[b]{0.47\textwidth}
  \centering
  \includegraphics[width=\textwidth]{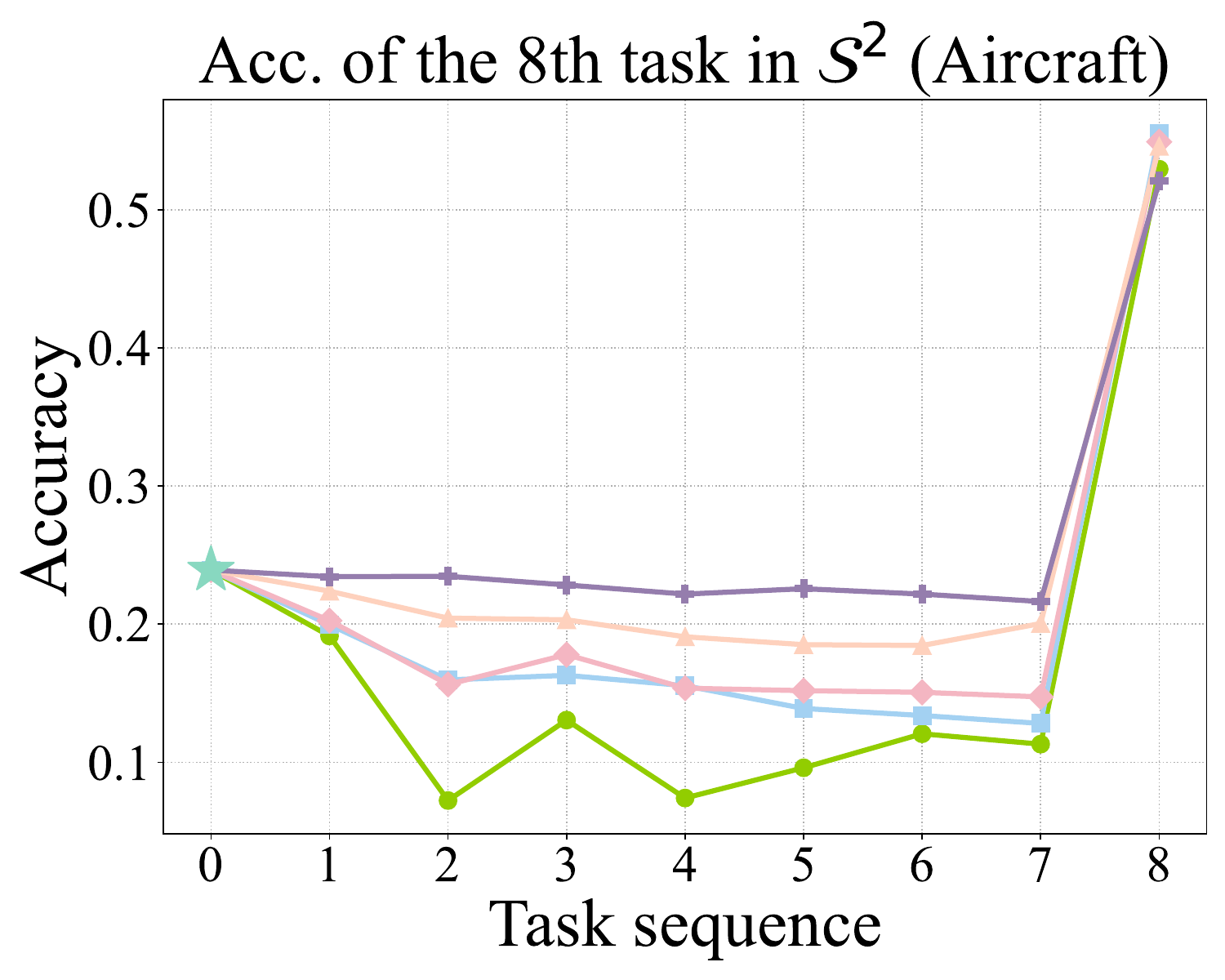}
\end{minipage}

\begin{minipage}[b]{0.47\textwidth}
  \centering
  \includegraphics[width=\textwidth]{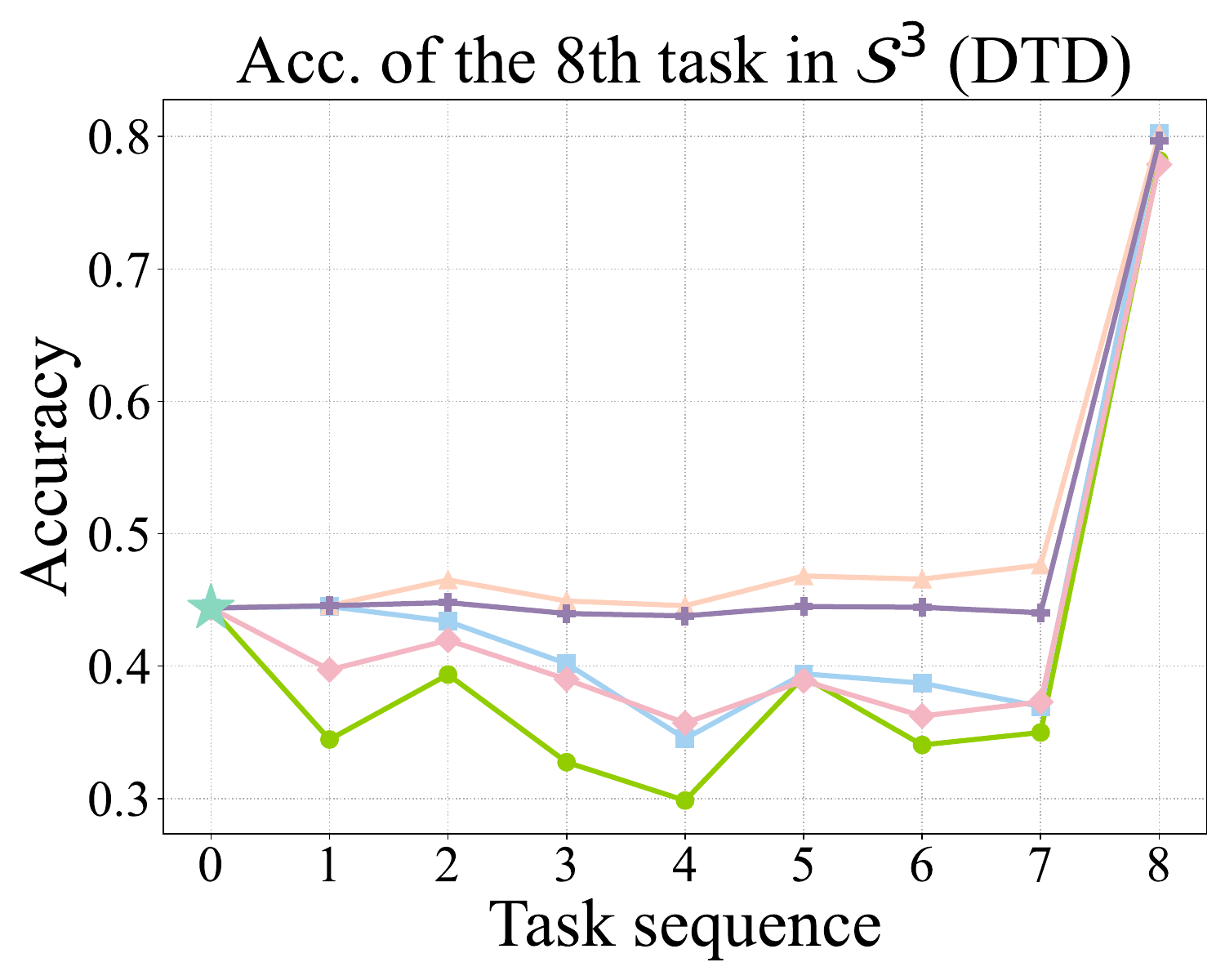}
\end{minipage}
\hspace{0.04\textwidth}
\begin{minipage}[b]{0.47\textwidth}
  \centering
  \includegraphics[width=\textwidth]{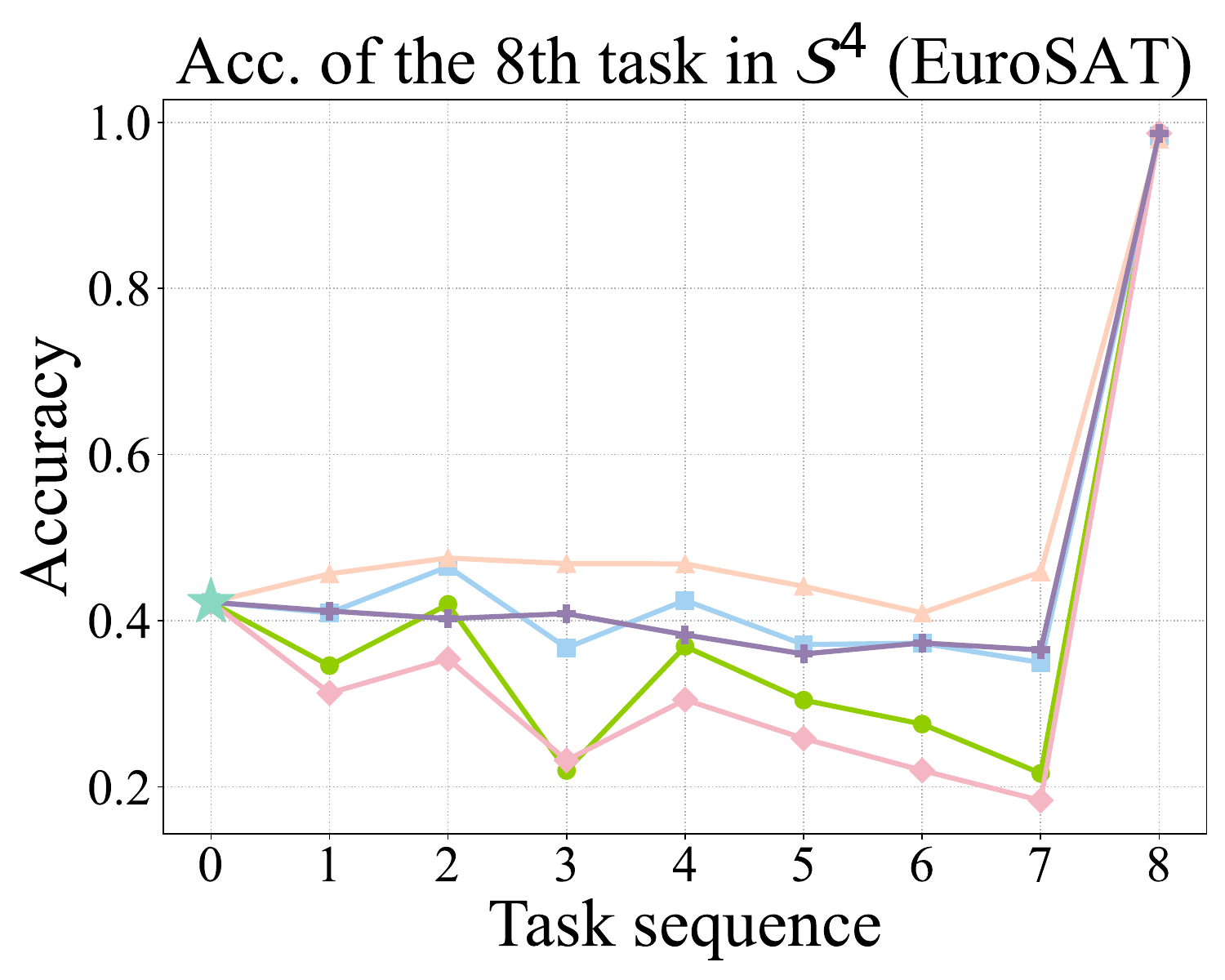}
\end{minipage}

\begin{minipage}[b]{0.47\textwidth}
  \centering
  \includegraphics[width=\textwidth]{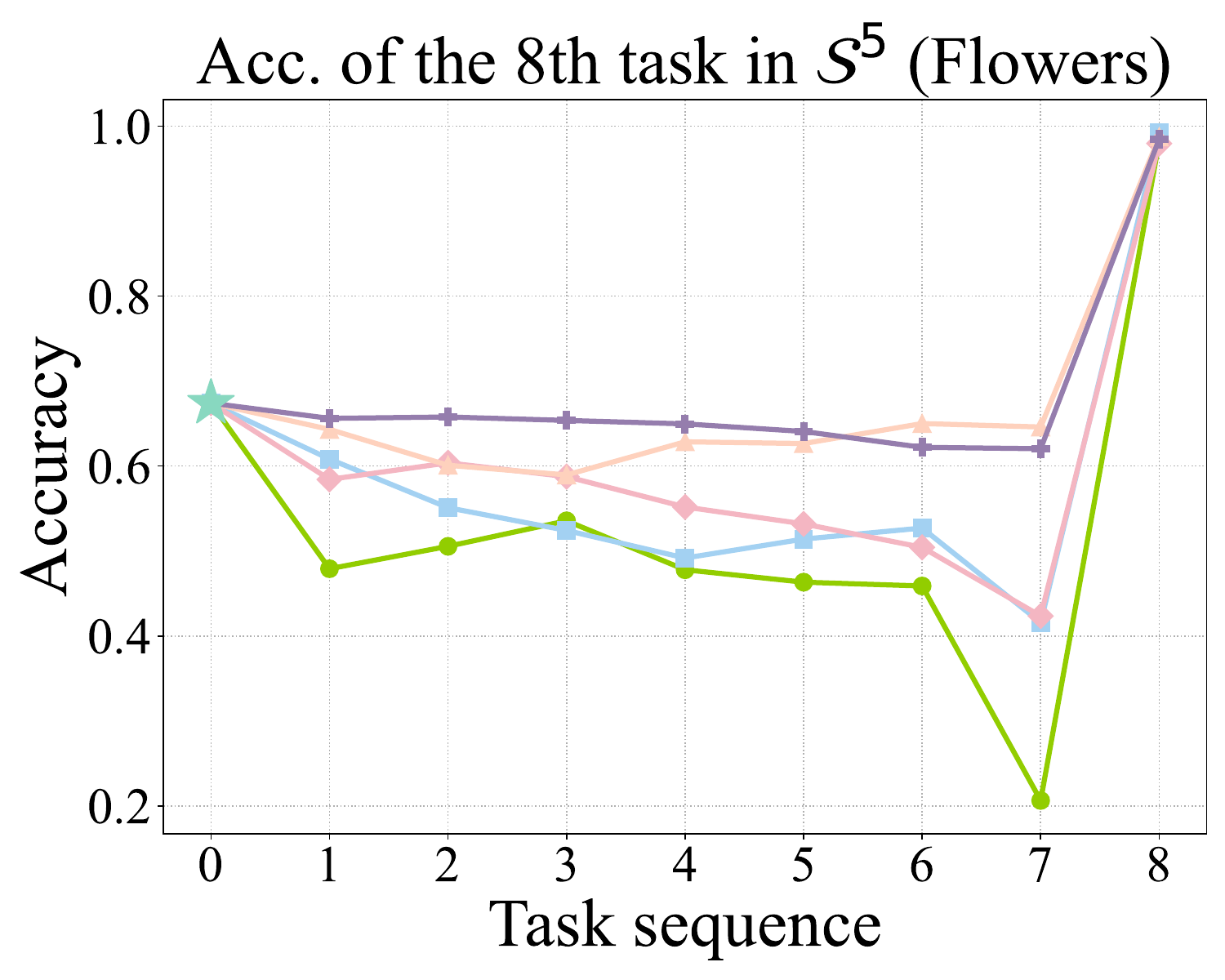}
\end{minipage}
\hspace{0.04\textwidth}
\begin{minipage}[b]{0.47\textwidth}
  \centering
  \includegraphics[width=\textwidth]{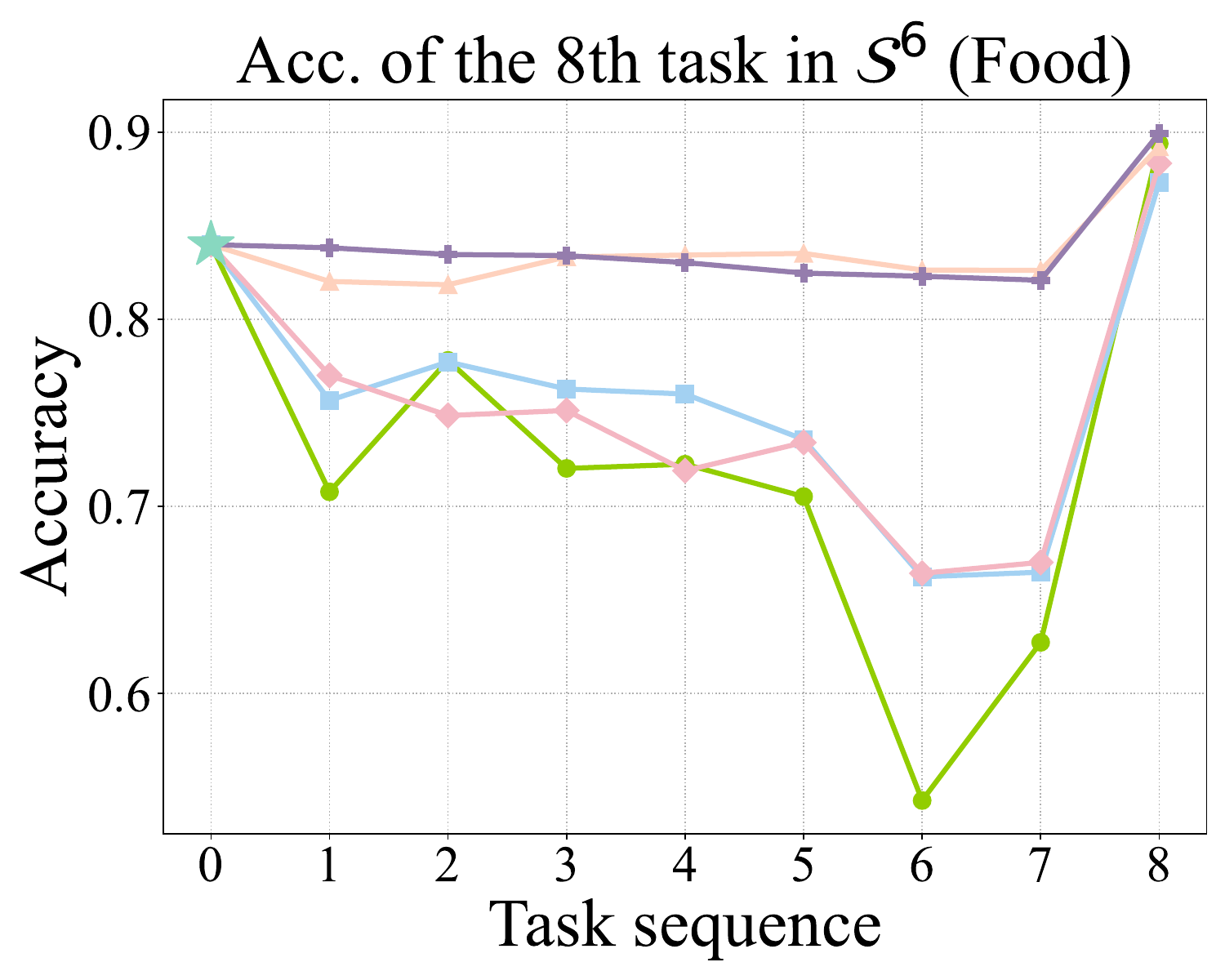}
\end{minipage}

\begin{minipage}[b]{0.47\textwidth}
  \centering
  \includegraphics[width=\textwidth]{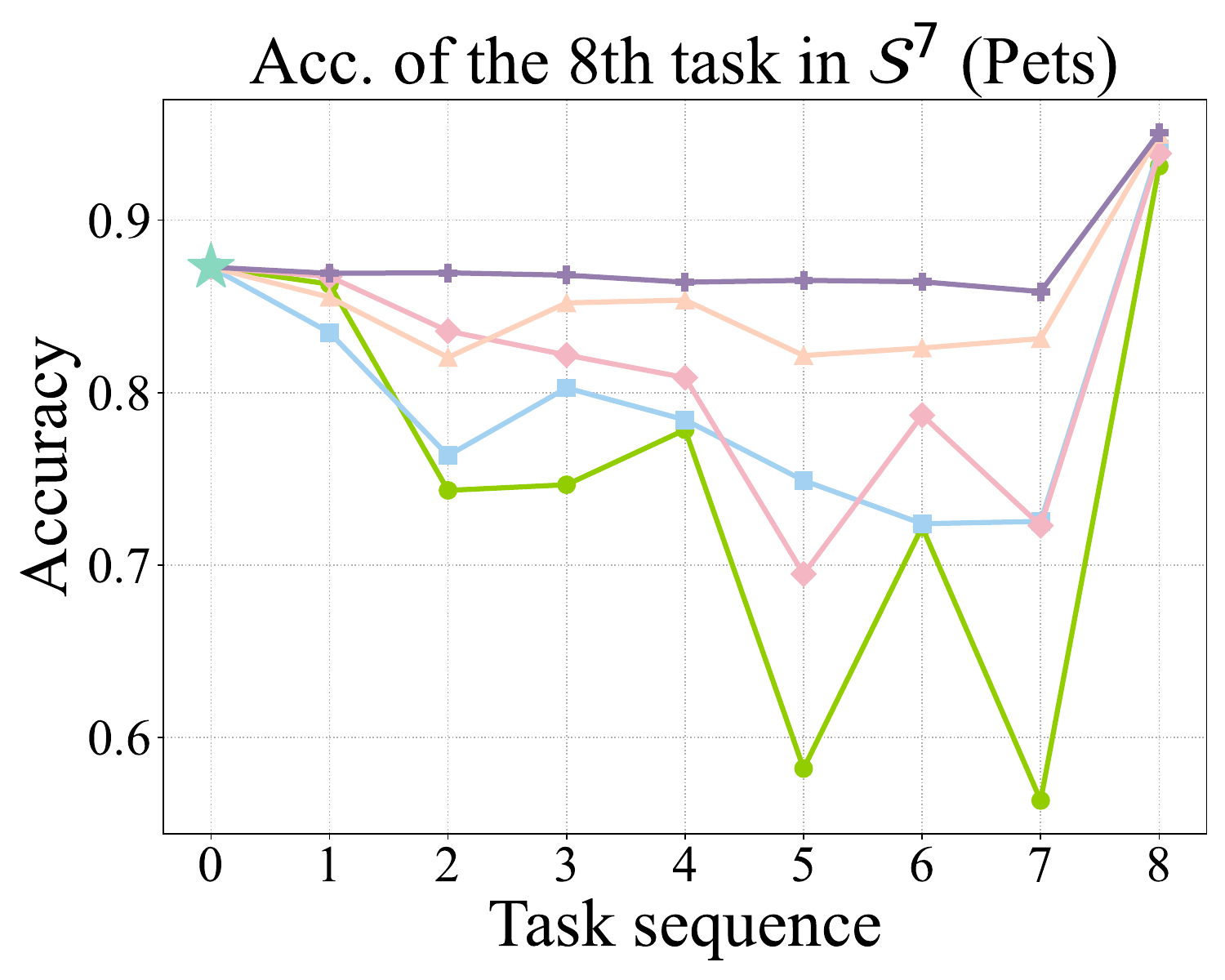}
\end{minipage}
\hspace{0.04\textwidth}
\begin{minipage}[b]{0.47\textwidth}
  \centering
  \includegraphics[width=\textwidth]{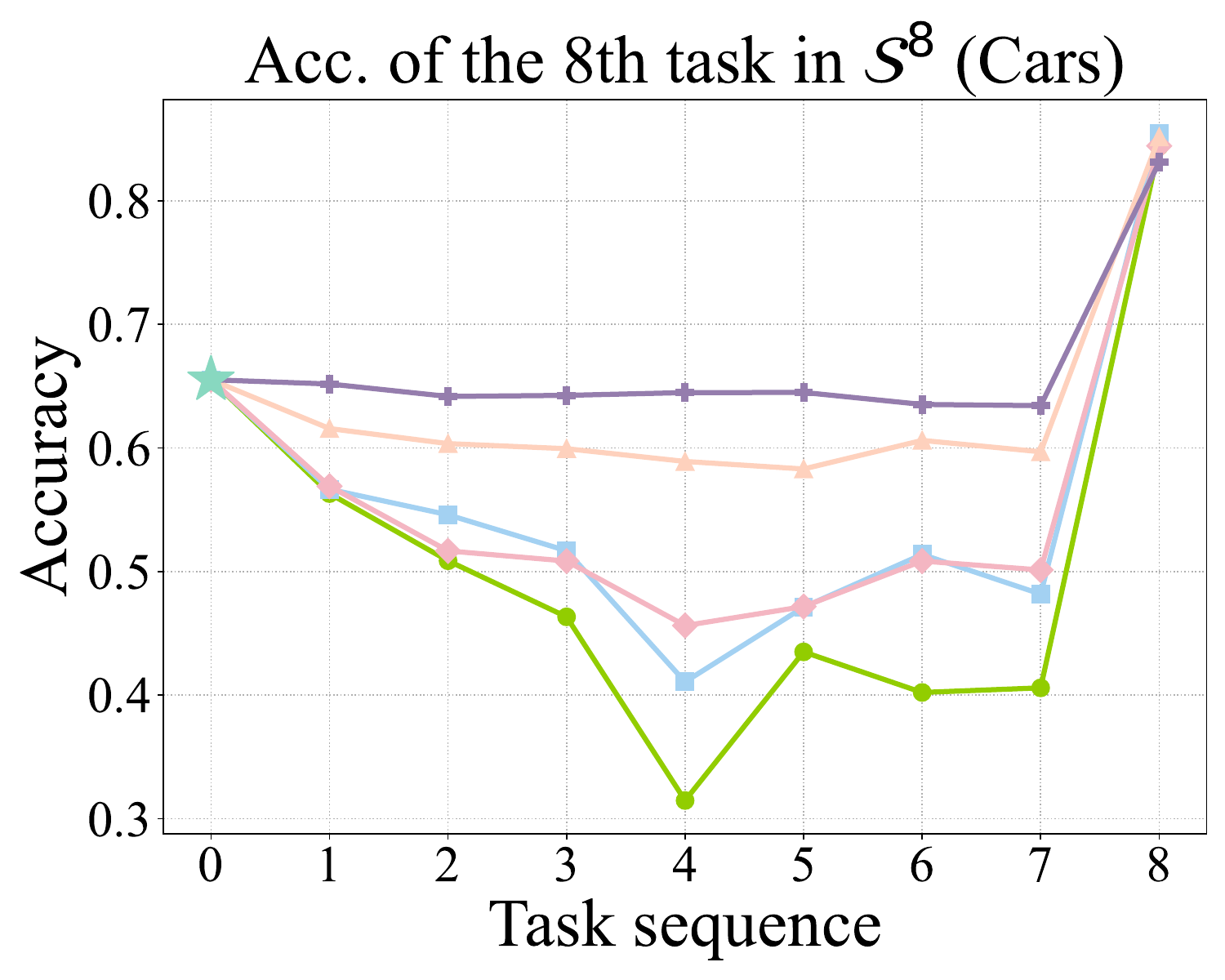}
\end{minipage}

\caption{Assessment of zero-shot degradation with the last task in the continual learning sequence (i.e., the horizontal axis). It can be seen that our method shows satisfactory accuracies before finetuning on the last task}

\label{fig:supp_degradation}
\end{figure}

\section{Experiments Details}

\subsubsection{Detailed Explanation to the Visualization of Reference Images with Large $\eta$ Scores.} 

To illustrate the reference images with the highest $\eta$ scores, we train our model on the first sequence $\calS^1$ (the detailed task orders are shown in \cref{tab:sequences}). For each stage $k \ge 2$, we calculate the $\eta$ scores for each reference image using \textbf{only} the original pre-trained model $g_0$ and the most recent fine-tuned model $g_{k-1}$ according to \cref{eq:similarity_estimation}. Then, we select the Top-25 images with the highest $\eta$ scores. Given that the visual concepts in some datasets are challenging to depict (\eg, EuroSAT, UCF101), we focus our visualizations on datasets with more concrete concepts, such as Flowers and Food, as visualized in \cref{fig:supp_visual}.



\subsubsection{Detailed results for Catastrophic Forgetting and Zero-Shot Degradation.}

In \cref{fig:forgetting} and \cref{fig:degradation}, we present examples of the assessment of catastrophic forgetting for the first task and evaluation of zero-shot degradation for the last task, respectively. Here we plot the impact of catastrophic forgetting on the first task and the impact of zero-shot degradation on the last task across each sequence in \cref{fig:supp_forgetting} and \cref{fig:supp_degradation}. For catastrophic forgetting, our method clearly outperform other methods by stably preserving the performance on the previously fine-tuned task (1st task in this case). Regarding the issue of zero-shot degradation, our method effectively maintains the original zero-shot capabilities in most scenarios, highlighting our success in preserving both pre-trained and previously fine-tuned knowledge across diverse datasets and various sequences. 

\begin{algorithm}[h]
\caption{Selective Dual-Teacher Knowledge Transfer}
\label{alg:algorithm}
\textbf{Input}: A pre-trained VLM $g_0$, hyper-parameters $\delta, \gamma, \lambda_{\text{dual}}$.\\
\textbf{Data}: A sequence of training tasks $\calS = (\calT^1, \cdots, \calT^K)$ and a reference dataset $\calX^{\text{ref}}$. \\
\textbf{Output}: The final fine-tuned model $g_K$.
\begin{algorithmic}[1] 
\FOR{$k$ in $1:K$}
    \STATE Freeze $g_0$ as the pre-trained knowledge teacher.
    \STATE Freeze $g_{k-1}$ as the previously fine-tuned knowledge teacher.
    \STATE Initialize the current model $g_k$ by $g_{k-1}$.
    \FOR{$e$ in $E$}
        \WHILE{not traverse over all current data $\calT^k$}
            \STATE Sample a batch of current data $B^k$.
            \STATE Sample a batch of ref data $B^{\text{ref}}$.
            \STATE Calculate $\calL_{\text{CE}}$ with the current data $B^k$.
            \STATE Calculate \cref{eq:dual} with $g_0$, $g_{k-1}$, and $B^{\text{ref}}$.
            \STATE Update $g_k$ with loss function \cref{eq:total}.
        \ENDWHILE
    \ENDFOR
\ENDFOR

\end{algorithmic}
\end{algorithm}
\section{The Training Algorithm of Our Proposed Framework}

As discussed in \cref{sec:training}, we provide the detailed training algorithm of our \emph{Selective Dual-Teacher Knowledge Transfer} framework in \cref{alg:algorithm}.

\end{document}